\documentclass[10pt,twocolumn,letterpaper]{article}

\makeatletter
\@namedef{ver@everyshi.sty}{}
\makeatother%

\usepackage{cvpr}              %

\usepackage{graphicx}
\usepackage{amsmath}
\usepackage{amssymb}
\usepackage{booktabs}
\usepackage{makecell}

\usepackage[accsupp]{axessibility}  %

\let\citep\cite
\let\citet\cite

\usepackage{amsmath,amsfonts,bm}

\def\eqref#1{equation~\ref{#1}}

\def\1{\bm{1}}

\def\vzero{{\bm{0}}}

\def\vx{{\bm{x}}}

\DeclareMathAlphabet{\mathsfit}{\encodingdefault}{\sfdefault}{m}{sl}
\SetMathAlphabet{\mathsfit}{bold}{\encodingdefault}{\sfdefault}{bx}{n}

\def\gN{{\mathcal{N}}}

\newcommand{\bc}{\mathbf{c}}

\newcommand{\bk}{\mathbf{k}}

\newcommand{\bx}{\mathbf{x}}

\newcommand{\btheta}{{\boldsymbol{\theta}}}

\newcommand{\bepsilon}{{\boldsymbol{\epsilon}}}

\usepackage{graphicx}
\usepackage{booktabs}
\usepackage{subcaption}
\usepackage{wrapfig}
\usepackage{comment}
\usepackage{float}
\usepackage{url}
\usepackage{graphicx}
\usepackage{comment}
\usepackage{amsmath,amssymb,amsthm}
\usepackage{bbm}
\usepackage{multirow}
\usepackage{caption}
\usepackage{booktabs}
\usepackage{siunitx}
\usepackage{comment}
\usepackage{amsmath,amssymb,amsthm}
\usepackage{pifont}%
\usepackage{makecell}
\usepackage{multirow}
\usepackage{times}
\usepackage{placeins}
\usepackage{colortbl}
\usepackage{rotating}
\usepackage{array}
\usepackage{color}
\usepackage{xcolor}
\usepackage{verbatim}
\usepackage{textcomp}
\usepackage{bbding}
\usepackage{mwe}
\usepackage{graphicx}
\usepackage{wrapfig}
\usepackage{subcaption}
\usepackage{listings}
\usepackage{bbm}
\usepackage{pifont}

\usepackage{booktabs}
\usepackage{chngcntr}
\usepackage{graphicx}
\usepackage{makecell}
\usepackage{mathtools}
\usepackage{microtype}
\usepackage{multirow}
\usepackage{nicefrac}
\usepackage{pgffor}
\usepackage{soul}
\usepackage{subcaption}
\usepackage{color, colortbl}
\usepackage{xr}

\definecolor{Gray}{gray}{0.9}
\definecolor{lightgray}{gray}{0.4}
\definecolor{verylightgray}{gray}{0.7}
\definecolor{veryverylightgray}{gray}{0.9}

\definecolor{darkgreen}{rgb}{0, 0.4, 0}
\definecolor{darkred}{rgb}{0.7, 0, 0}
\definecolor{darkblue}{rgb}{0.0, 0.0, 0.7}

\newcommand{\cmark}{\textcolor{black}{\ding{51}}}%
\newcommand{\xmark}{\textcolor{black}{\ding{55}}}%

\newcommand{\ttt}[1]{{#1}}

\usepackage[pagebackref,breaklinks,colorlinks]{hyperref}

\usepackage[capitalize]{cleveref}
\crefname{section}{Sec.}{Secs.}
\Crefname{section}{Section}{Sections}
\Crefname{table}{Table}{Tables}
\crefname{table}{Tab.}{Tabs.}

\newcommand\blfootnote[1]{%
  \begingroup
  \renewcommand\thefootnote{}\footnote{#1}%
  \addtocounter{footnote}{-1}%
  \endgroup
}

\newcommand{\customfootnotetext}[2]{{%
  \renewcommand{\thefootnote}{#1}%
  \footnotetext[0]{#2}}}%

\def\cvprPaperID{621} %
\def\confName{CVPR}
\def\confYear{2023}

\begin{document}

\title{Self-Guided Diffusion Models}

\author{
Vincent Tao Hu\textsuperscript{1$\dagger$} \hspace{2mm}
David W. Zhang\textsuperscript{1$\dagger$} \hspace{2mm}
\\
Yuki M. Asano\textsuperscript{1} 
\hspace{2mm}
Gertjan J. Burghouts\textsuperscript{2} \hspace{2mm} 
Cees G. M. Snoek\textsuperscript{1}  \\
\vspace{1mm}
\textsuperscript{1}University of Amsterdam \qquad \textsuperscript{2}TNO \\ 
}
\maketitle

\customfootnotetext{${\dagger}$}{Equal contribution, taohu620@gmail.com}
\begin{abstract}
   Diffusion models have demonstrated remarkable progress in image generation quality, especially when guidance is used to control the generative process. However, guidance requires a large amount of image-annotation pairs for training and is thus dependent on their availability and correctness. In this paper, we eliminate the need for such annotation by instead exploiting the flexibility of self-supervision signals to design a framework for \textit{self-guided} diffusion models. By leveraging a feature extraction function and a self-annotation function, our method provides guidance signals at various image granularities: from the level of holistic images to object boxes and even segmentation masks. Our experiments on single-label and multi-label image datasets demonstrate that self-labeled guidance always outperforms diffusion models without guidance and may even surpass guidance based on ground-truth labels. When equipped with self-supervised box or mask proposals, our method further generates visually diverse yet semantically consistent images, without the need for any class, box, or segment label annotation. Self-guided diffusion is simple, flexible and expected to profit from deployment at scale.
\blfootnote{Source code  will be at: \url{https://taohu.me/sgdm/}.}

\end{abstract}

\begin{figure}[t]
    \centering
    \vspace{1mm}
    \includegraphics[width=0.48\textwidth]{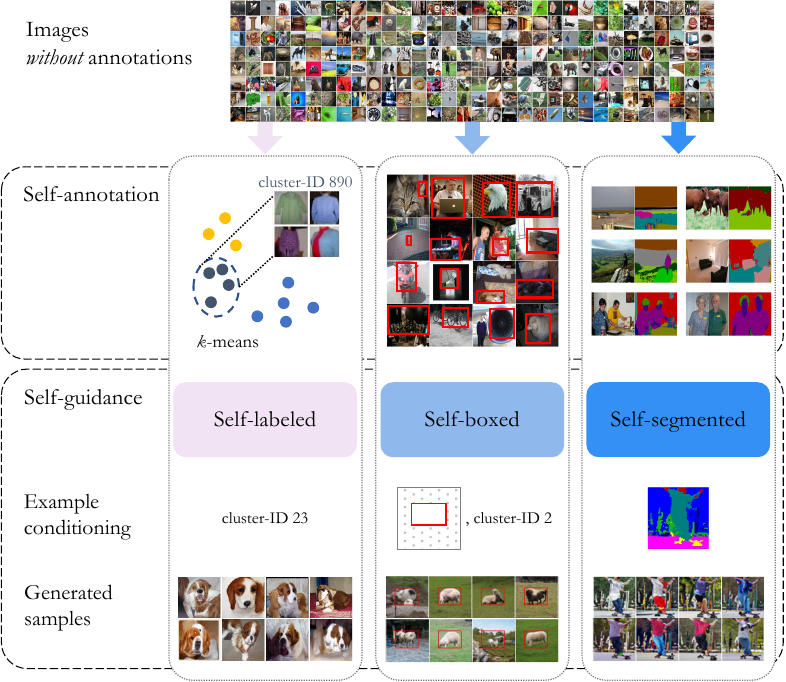}
    \caption{
    \textbf{Self-guided diffusion framework.}
    Our method can leverage large and diverse image datasets \textit{without} any annotations for training guided diffusion models. 
    Starting from a dataset without ground-truth annotations, we apply a self-supervised feature extractor to create self-annotations. Using these, we train diffusion models with either self-labeled, self-boxed, or self-segmented guidance that enable controlled generation and improved image fidelity.
    }
    \label{fig:pipeline}
\end{figure}

\section{Introduction}
Diffusion models have recently enabled tremendous advancements in many computer vision fields related to image synthesis, but counterintuitively this often comes with the cost of requiring large annotated datasets~\citep{saharia2022photorealistic_imagen, ramesh2022hierarchical_dalle2}.
For example, the image fidelity of samples from diffusion models can be spectacularly enhanced by conditioning on class labels~\citep{dhariwal2021diffusion_beat}. Classifier guidance goes a step further and offers control over the alignment with the class label, by using the classifier gradient to guide the image generation~\citep{dhariwal2021diffusion_beat}. Classifier-free guidance~\citep{ho2021classifier} replaces the dedicated classifier with a diffusion model trained by randomly dropping the condition during training. This has proven a fruitful line of research for several other condition modalities, such as text~\citep{saharia2022photorealistic_imagen,ramesh2021zero_dalle1}, image layout~\citep{rombach2022high_latentdiffusion_ldm}, visual neighbors~\citep{ashual2022knn}, and image features~\citep{giannone2022_fewshot}. However, all these conditioning and guidance methods require ground-truth annotations. 
In many domains, this is an unrealistic and too costly assumption. For example, medical images require domain experts to annotate very high-resolution data, which is infeasible to do exhaustively~\citep{panteli2021sparse}. In this paper, we propose to remove the necessity of ground-truth annotation for guided diffusion models.

We are inspired by progress in self-supervised learning \citep{chen2020big_simclr_v2,dino}, which encodes images into semantically meaningful latent vectors without using any label information. It usually does so by solving a pretext task~\citep{zhang2017split,gidaris2018unsupervised_rotate,asano2019selflabelling,he2020momentum_moco} on image-level to remove the necessity of labels. This annotation-free paradigm enables the representation learning to upscale to larger and more diverse image datasets~\citep{gao2021large}. The holistic image-level self-supervision has recently been extended to more expressive dense representations, including bounding boxes (e.g.,~\citep{simeoni2021localizing_lost, melas2022deep}) and pixel-precise segmentation masks (e.g.,~\citep{hamilton2022unsupervised_stego,ziegler2022self}). Some self-supervised learning methods even outperform supervised alternatives~\citep{he2020momentum_moco,dino}. We hypothesize that for diffusion models, self-supervision may also provide a flexible and competitive, possibly even stronger guidance signal than ground-truth labeled guidance. 

In this paper, we propose \textit{self-guided diffusion models}, a framework for image generation using guided diffusion without the need for any annotated image-label pairs, the detailed structure is shown in ~\Cref{fig:pipeline}. The framework encompasses a feature extraction function and a self-annotation function, that are compatible with recent self-supervised learning advances. Furthermore, we leverage the flexibility of self-supervised learning to generalize the guidance signal from the holistic image level to (unsupervised) local bounding boxes and segmentation masks for more fine-grained guidance. We demonstrate the potential of our proposal on single-label and multi-label image datasets, where self-labeled guidance always outperforms diffusion models without guidance and may even surpass guidance based on ground-truth labels. When equipped with self-supervised box or mask proposals, our method further generates visually diverse yet semantically consistent images, without the need for any class, box, or segment label annotation.

\section{Related Work}

\paragraph{Conditional generative models.} 
Earlier works on generative adversarial networks (GANs) have observed improvements in image quality by conditioning on ground-truth labels~\citep{mirza2014conditional_cgan,brock2018large_biggan,icgan}.
Recently, conditional diffusion models have reported similar improvements, while also offering a great amount of controllability via classifier-free guidance by training on images paired with textual descriptions~\citep{ramesh2021zero_dalle1,ramesh2022hierarchical_dalle2,saharia2022photorealistic_imagen}, semantic segmentations~\citep{wang2022semantic}, or other modalities~\citep{bordes2021high,yang2022diffusion,song2021solving}.
Our work also aims to realize the benefits of conditioning and guidance, but instead of relying on additional human-generated supervision signals, we leverage the strength of pretrained self-supervised visual encoders.

Zhou \etal \citet{zhou2021lafite} train a GAN for text-to-image generation without any image-text pairs, by leveraging the CLIP~\citep{radford2021learning_clip} model that was pretrained on a large collection of paired data. In this work, we do not assume any paired data for the generative models and rely purely on images. Additionally, image layouts are difficult to be expressed by text, thus our self-boxed and self-segmented methods are complementary to text conditioning.  
Instance-Conditioned GAN~\citep{icgan}, Retrieval-augmented Diffusion~\citep{blattmann2022retrieval} and KNN-diffusion~\citep{ashual2022knn} are three recent methods that utilize nearest neighbors as guidance signals in generative models.
Similar to our work, these methods rely on conditional guidance from an unsupervised source, 
we differ from them by further attempting to provide more diverse \textit{spatial} guidance, including (self-supervised) bounding boxes and segmentation masks.

\paragraph{Self-supervised learning in generative models.} 
Self-supervised learning~\citep{caron2020unsupervised_swap,chen2020big_simclr_v2,asano2019selflabelling,dino} has shown great potential for representation learning in many downstream tasks. 
As a consequence, it is also  commonly explored in GAN for evaluation and analysis~\citep{morozov2020self}, conditioning~\citep{icgan,mangla2022data}, stabilizing training~\citep{chen2019self}, reducing labeling costs~\citep{luvcic2019high} and avoiding mode collapse~\citep{armandpour2021partition}.
Our work focuses on translating the benefits of self-supervised methods to the generative domain and providing flexible guidance signals to diffusion models at various image granularities. 
In order to analyze the feature representation from self-supervised models, Bordes \etal \citet{bordes2021high} condition on self-supervised features in their diffusion model for better visualization in data space. We instead condition on the compact clustering after the self-supervised feature, and further introduce the elasticity of self-supervised learning into diffusion models for multi-granular image generation.

\section{Approach}
Before detailing our self-guided diffusion framework, we provide a brief background on diffusion models %
and the classifier-free guidance technique.

\subsection{Background}
\label{sec:background}

\paragraph{Diffusion models.}
Diffusion models~\cite{sohl2015deep,ho2020denoising} gradually add noise to an image $\bx_0$ until the original signal is fully diminished. By learning to reverse this process one can turn random noise $\bx_T$ into images. This diffusion process is modeled as a Gaussian process with Markovian structure:
\begin{equation}\label{eq:forward}
\begin{split}
q(\bx_t|\bx_{t-1})&\coloneqq\mathcal{N}(\bx_t;\sqrt{1-\beta _t}\bx_{t-1},\beta_{t}\mathbf{I}) \\
q(\bx_{t}|\bx_{0})&\coloneqq\mathcal{N}(\bx_t;\sqrt{\overline{\alpha}_t}\bx_{0},(1-\overline{\alpha}_t)\mathbf{I},
\end{split}
\end{equation}
where $\beta_{1},\ldots,\beta_{T}$ is a fixed variance schedule on which we define $\alpha _{t}\coloneqq1-\beta _{t}$ and $\overline{\alpha}_{t}\coloneqq\prod_{s=1}^t \alpha _{s}$. All latent variables have the same dimensionality as the image $\bx_0$ and differ by the proportion of the retained signal and added noise.

Learning the reverse process reduces to learning a denoiser $\bx_t\sim q(\bx_t|\bx_0)$ that recovers the original image as  $(\bx_t - (1-\overline{\alpha}_{t})\bepsilon_{\btheta}(\bx_t, t))/\sqrt{\overline{\alpha}_{t}}\approx \bx_0$.
Ho \etal~\cite{ho2020denoising} optimize the parameters $\btheta$ of noise prediction network by minimizing:
\begin{equation}\label{eq:loss_fn}
    \mathcal{L}(\btheta) = \mathbb{E}_{\bepsilon,\bx, t}\left[||\bepsilon_{\btheta}(\bx_{t},t) - \bepsilon||^{2}_{2}\right],
\end{equation}
in which $\bepsilon \sim \mathcal{N}(\vzero,\mathbf{I})$, $\bx \in \mathcal{D}$ is a sample from the training dataset $\mathcal{D}$ and  the noise prediction function $\bepsilon_{\theta}(\cdot)$ are encouraged to be as close as possible to $\bepsilon$.

The standard %
sampling~\cite{ho2020denoising} requires many neural function evaluations to get good quality samples. Instead, the faster Denoising Diffusion Implicit Models (DDIM) sampler~\cite{song2020denoising_ddim} 
has a non-Markovian sampling process:
\begin{multline}
    \bx_{t-1} = \sqrt{\overline{\alpha}_{t-1}} \left(\frac{\bx_t - \sqrt{1 - \overline{\alpha}_t} \bepsilon_\theta(\bx_t,t)}{\sqrt{\overline{\alpha}_t}}\right) \\
    + \sqrt{1 - \overline{\alpha}_{t-1} - \sigma_t^2} \cdot \bepsilon_\theta(\bx_t,t) + \sigma_t \bepsilon, \label{eq:sample-eq-gen}
\end{multline}
where $\bepsilon \sim \gN(\vzero, \mathbf{I})$ is Gaussian noise independent of $\vx_t$.

\paragraph{Classifier-free guidance.}  
To trade off mode coverage and sample fidelity in a conditional diffusion model, Dhariwal and Nichol~\citet{dhariwal2021diffusion_beat} propose to guide the image generation process using the gradients of a classifier, with the additional cost of having to train the classifier on noisy images. 
Motivated by this drawback, Ho and Salimans \cite{ho2021classifier} introduce label-conditioned guidance that does not require a classifier. They obtain a combination of a conditional and  unconditional network in a single model, by randomly dropping the guidance signal $\bc$ during training. After training, it empowers the model with progressive control over the degree of alignment between the guidance signal and the sample by varying the guidance strength $w$:
\begin{equation}\label{eq:cfg_default}
    \tilde{\bepsilon}_{\theta}(\bx_{t},t;~\bc, w) = (1-w) \bepsilon_{\theta}(\bx_{t},t) + w \bepsilon_{\theta}(\bx_{t},t;~\bc).
\end{equation}
A larger $w$ leads to greater alignment with the guidance signal, and vice versa. Classifier-free guidance~\cite{ho2021classifier} provides progressive control over the specific guidance direction at the expense of labor-consuming data annotation. In this paper, we propose to remove the necessity of data annotation using a self-guided principle based on self-supervised learning.

\subsection{Self-Guided Diffusion Models}
The equations describing the diffusion model for classifier-free guidance implicitly assume dataset $~\mathcal{D}$ and its images each come with a single manually annotated class label. We prefer to make the label requirement explicit. We denote the human annotation process as the function $\xi(\bx; \mathcal{D}, \mathcal{C}): \mathcal{D}\rightarrow\mathcal{C}$, where $\mathcal{C}$ defines the annotation taxonomy, and plug this into~\Cref{eq:cfg_default}:
\begin{multline}
    \tilde{\bepsilon}_{\theta}(\bx_{t},t;~\xi(\bx;~\mathcal{D},\mathcal{C}),w) = \\(1-w) \bepsilon_{\theta}(\bx_{t},t) + w \bepsilon_{\theta}(\bx_{t},t;~\xi(\bx;~\mathcal{D},\mathcal{C})).
\end{multline}
We propose to replace the supervised labeling process $\xi$ with a self-supervised process that requires \textit{no} human annotation:
\begin{multline}\label{eq:self_guidance}
    \tilde{\bepsilon}_{\theta}(\bx_{t},t; f_{\psi}(g_{\phi}(\bx;~\mathcal{D});~\mathcal{D}),w) = \\ (1-w) \bepsilon_{\theta}(\bx_{t},t) + w \bepsilon_{\theta}(\bx_{t},t;~f_{\psi}(g_{\phi}( \bx;~\mathcal{D});~\mathcal{D}),
\end{multline}
where $g$ is a  self-supervised feature extraction function parameterized by $\phi$ that maps the input data to feature space $\mathcal{H}$, $g:\bx \rightarrow g_{\phi}(\bx), \forall \bx \in \mathcal{D}$, and $f$ is a self-annotation function parameterized by $\psi$ to map the raw feature representation to the ultimate guidance signal $\bk$, $f_{\psi}: g_{\phi}(\cdot;\mathcal{D}) \rightarrow \bk$.
The guidance signal $\bk$ can be any form of \textit{annotation}, e.g., label, box, pixel, that can be paired with an image, which we derive by $\bk {=} f_{\psi}(g_{\phi}(\bx;~\mathcal{D});~\mathcal{D})$.
The choice of the self-annotation function $f$ can be non-parametric by heuristically searching over dataset~$\mathcal{D}$ based on the extracted feature $g_{\phi}(\cdot;~\mathcal{D})$, or parametric by fine-tuning  on the feature map $g_{\phi}(\cdot;~\mathcal{D})$. 

For the noise prediction function $\bepsilon_{\theta}(\cdot)$, we adopt the traditional UNet network architecture~\cite{unet} due to its superior image generation performance, following~\cite{ho2020denoising,song2021scorebased_sde,ramesh2022hierarchical_dalle2,saharia2022photorealistic_imagen}.

Stemming from this general framework, we present three  methods working at different spatial granularities, all without relying on any ground-truth labels.
Specifically, we cover image-level, box-level, and pixel-level guidance by setting the feature extraction function $g_{\phi}(\cdot)$, self-annotation function $f_{\psi}(\cdot)$, and guidance signal $\bk$ to an approximate form.

\paragraph{Self-labeled guidance.} 
To achieve self-labeled guidance, we need a self-annotation function $f$ that produces a  representative guidance signal $\bk \in \mathbb{R}^{K}$. 
Firstly, we need an embedding function  $g_{\phi}(\bx), \bx \in \mathcal{D}$ which  provides semantically meaningful  image-level guidance for the model. We obtain $g_{\phi}(\cdot)$ in a self-supervised manner by mapping from image space, $g_{\phi}(\cdot): \mathbb{R}^{W\times H \times 3} \rightarrow \mathbb{R}^{C}$, where $W$ and $H$ are image width and height and $C$ is the feature dimension. We may use any type of feature for the feature embedding function $g$, which we will vary and validate in the experiments. 
As the image-level feature $g_\phi(\cdot;~\mathcal{D})$ is not compact enough for guidance,  we further conduct a non-parametric clustering algorithm, e.g., $k$-means, as our self-annotation function $f$. For all features $g_{\phi}(\cdot)$, we  
obtain the self-labeled guidance via self-annotation function $f_\psi(\cdot): \mathbb{R}^{C} \rightarrow \mathbb{R}^{K}$. Motivated by~\citet{rolfe2016discrete}, we use a one-hot embedding  $\bk \in \mathbb{R}^{K}$ for each image to achieve a compact guidance.%

We inject the guidance information into the noise prediction function $\bepsilon_{\theta}$ by concatenating it with timestep embedding $t$  and feed the concatenated information \texttt{concat}$[t, \bk]$ into every block of the UNet. Thus, the noise prediction function $\bepsilon_{\theta}$ is rewritten as: %
\begin{equation}\label{eq:sampling_labelfree}
\bepsilon _{\theta }(\bx_{t},t;~\bk) = \bepsilon_{\theta }(\bx_{t},\texttt{concat}[t,~\bk]),
\end{equation}
where  $\bk {=} f_{\psi}(g_{\phi}(\bx;~\mathcal{D});~\mathcal{D})$ is the self-annotated image-level guidance signal.
For simplicity, we ignore the self-annotation function $f_{\psi}(\cdot)$ here and in the later text. 
Self-labeled guidance focuses on image-level global guidance. Next, we consider a more fine-grained spatial guidance.

\paragraph{Self-boxed guidance.}
Bounding boxes specify the location of an object in an image~\cite{ren2015faster,carion2020end_detr} and complement the content information provided by class labels.
Our self-boxed guidance approach aims to attain this signal via self-supervised models. We represent the bounding box as a binary mask $\bk_s\in\mathbb{R}^{W \times H}$ rather than coordinates, where 1 indicates that the pixel is inside the box and 0 outside. This design directly aligns the image and mask along the spatial dimensions.
We propose the  self-annotation function $f$ that obtains bounding box $\bk_{s}$  by mapping from feature space~$\mathcal{H}$ to the bounding box space via $f_{\psi}(\cdot;~\mathcal{D}): \mathbb{R}^{W \times H \times C}  \rightarrow \mathbb{R}^{W \times H}$, and inject the guidance signal by concatenating in the channel dimension:  $\bx_{t}\coloneqq\texttt{concat}[\bx_{t}$, $\bk_{s}]$.  %
Usually in self-supervised learning, the derived bounding box is class-agnostic~\cite{vo2020toward,vo2021large}. 
To inject a self-supervised pseudo label to further enhance the guidance signal, we again resort to  clustering to obtain  $\bk$ and concatenate it with the time embedding $t\coloneqq\texttt{concat}[t, \bk]$.
To incorporate such guidance, we reformulate the noise prediction function $\bepsilon_\theta$ as:
\begin{multline}\label{eq:sampling_box}
\bepsilon _{\theta }(\bx_{t},t;~\bk_{s},\bk) = \\ \bepsilon _{\theta }(\texttt{concat}[\bx_{t},\bk_{s}],\texttt{concat}[t, \bk]),\qquad
\end{multline}
in which $\bk_{s}$ is the self-supervised box guidance obtained by self-annotation functions $f_{\psi}$, $\bk$ is the self-supervised image-level guidance from clustering. %
$\bk_{s}$ and $\bk$ denotes the location and class information, respectively.
The design of $f_{\psi}$ is flexible as long as it obtains self-supervised bounding boxes by $f_{\psi}(\cdot;~\mathcal{D}): \mathbb{R}^{W \times H \times C}  \rightarrow \mathbb{R}^{W \times H }$. Self-boxed guidance  guides the diffusion model by boxes, which specifies the box area in which the object will be generated. Sometimes, we may need an even finer granularity, e.g., pixels, which we detail next.

\paragraph{Self-segmented guidance.}
Compared to a bounding box, a segmentation mask is a more fine-grained signal. Additionally, a multichannel mask is more expressive than a binary foreground-background mask. Therefore, we propose a self-annotation function $f$ that acts as a plug-in built on feature $g_{\phi}(\cdot;~\mathcal{D})$ to extract the segmentation mask $\bk_{s}$ via function mapping $ f _{\psi}(\cdot;~\mathcal{D}): \mathbb{R}^{W \times H \times C} \rightarrow \mathbb{R}^{W \times H \times K}$, where $K$ is the number of segmentation clusters. 
 
To inject the self-segmented guidance into the noise prediction function $\bepsilon_\theta $, we consider two pathways for injection of such guidance. We first concatenate the segmentation mask to $\bx_{t}$ in the channel dimension, $\bx_{t}\coloneqq\texttt{concat}[\bx_{t}$, $\bk_{s}]$, to retain the spatial inductive bias of the guidance signal.
Secondly, we also incorporate the image-level guidance 
to further amplify the guidance signal along the channel dimension. As the segmentation mask from the self-annotation function $f_{\psi}$ already contains  image-level information, we do not apply the image-level clustering as before in our self-labeled guidance. Instead, we directly derive the image-level guidance from the self-annotation result $f_{\psi}(\cdot)$  via spatial maximum pooling: $\mathbb{R}^{W \times H \times K} \rightarrow \mathbb{R}^{K}$, 
and feed the  image-level guidance $\hat{\bk}$ into the noise prediction function via concatenating it with the timestep embedding $t\coloneqq$\texttt{concat}$[t, \hat{\bk}]$. The concatenated results will be sent to every block of the UNet. In the end, the overall noise prediction function for self-segmented guidance is formulated as:
\begin{multline}\label{eq:sampling_mask}
\bepsilon _{\theta }(\bx_{t},t;~\bk_{s},\hat{\bk}) = \\ \bepsilon _{\theta }(\texttt{concat}[\bx_{t},\bk_{s}],~\texttt{concat}[t, \hat{\bk}]),\qquad
\end{multline}
in which  $\bk_{s}$ is the spatial mask guidance obtained from self-annotation function $f$, $\hat{\bk}$ is a multi-hot image-level guidance derived from the self-supervised learning mask $\bk_{s}$.

We have described three variants of self-guidances by setting the feature extraction function $g_{\phi}(\cdot)$, self-annotation function $f_{\psi}(\cdot)$, guidance signal $\bk$ to an approximate form. In the end, we arrive at three noise prediction functions $\bepsilon_{\theta}$, which we utilize for diffusion model training and sampling, following the standard  guided \cite{ho2021classifier} diffusion approach as detailed in~\Cref{sec:background}.

\section{Experiments}

In this section, we aim to answer the overarching question: Can we substitute ground-truth annotations with self-annotations? First, we consider the image-label setting, in which we examine what kind of self-labeling is required to improve image fidelity. 
In addition, we explore what semantic concepts are induced by self-labeling approaches that broaden the control over the content beyond the standard ground-truth labels. 
Next, we look at image-bounding box pairs. Finally, we examine whether it is possible to gain fine-grained control with self-labeled image-segmentation pairs. We first present the general settings relevant for all experiments.

\paragraph{Evaluation metric.} 
We evaluate both diversity and fidelity of the generated images by the Fr\'echet Inception Distance (FID)~\citep{heusel2017gans_fid}, as it is the de facto metric for the evaluation of generative methods, e.g.,~\citep{dhariwal2021diffusion_beat,karras2019_stylegan,brock2018large_biggan,saharia2022photorealistic_imagen}. It provides a symmetric measure of the distance between two distributions in the feature space of Inception-V3~\citep{szegedy2016rethinking_inceptionv3}. We use FID as our main metric for the sampling quality.

\paragraph{Baselines \& implementation details.}
As baselines, we compare against both the unconditional diffusion model and a diffusion model trained with classifier-free guidance using ground-truth annotations~\citep{ho2021classifier}.
We use the same neural network and hyperparameters for the baselines and our method.
Note that applying more training steps generally tends to further improve the performance~\citep{kingma2021variational, saharia2022photorealistic_imagen}, thus to facilitate a fair comparison we use the same computational budget in every experiment when comparing the baselines to our proposed method. We use DDIM~\citep{song2020denoising_ddim} samplers with 250 steps, $\sigma_t{=}0$ to efficiently generate samples. For details on the hyperparameters, we refer to~\Cref{sec:hyperparames}.

\subsection{Self-Labeled Guidance}

We use ImageNet32/64~\citep{deng2009imagenet} and CIFAR100~\citep{krizhevsky2009learning} to validate the efficacy of self-labeled guidance. On ImageNet, we also measure the Inception Score (IS)~\citep{salimans2016improved_is_inceptionscore}, following common practice~\citep{dhariwal2021diffusion_beat,karras2019_stylegan,brock2018large_biggan}. IS measures how well a model fits into the full ImageNet class distribution.

\begin{table}%
\centering
\resizebox{0.63\linewidth}{!}{
\begin{tabular}{lrrr}
\toprule
 &  FID$\downarrow$ & IS$\uparrow$ \\
 \midrule
 \rowcolor{Gray}
\textbf{Label-supervised} & & &\\
 ResNet50  & 22.00  & 8.23 \\ 
 ViT-B/16   & 22.30 & 7.81 \\ 
 \midrule
 \rowcolor{Gray}
 \textbf{Self-supervised} & & &\\
 MAE ViTBase  & 32.58 & 8.20 \\
 SimCLR-v2  & \ttt{23.16} & 9.35 \\ 
 MSN ViT-B/16 & 21.16 & \textbf{10.59}\\ 
 DINO ViT-B/16 & \textbf{19.35} & 10.41 \\
\bottomrule
\end{tabular}}
\caption{ \textbf{Choice of feature extraction function} on ImageNet32. DINO and MSN \texttt{ViT-B/16} obtain good trade-offs between FID and IS. }

\label{tab:backone}
\end{table}

\paragraph{Choice of feature extraction function $\bm{g}$.} We first measure the influence of the feature extraction function $g$ used before clustering. We consider two supervised feature backbones: \texttt{ResNet50}~\citep{he2016deep_resnet} and \texttt{ViT-B/16}~\citep{dosovitskiy2020image_vit}, and four self-supervised backbones: \texttt{SimCLR}~\citep{chen2020big_simclr_v2}, \texttt{MAE}~\citep{he2022masked_mae}, \texttt{MSN}~\citep{assran2022masked_msn} and \texttt{DINO}~\citep{dino}. To assure a fair comparison we use 10k clusters for all architectures.
From the results in~\Cref{tab:backone}, we make the following observations. First, features from the supervised \texttt{ResNet50}, and \texttt{ViT-B/16} lead to a satisfactory FID performance, at the expense of relatively limited diversity (low IS). However, they still require label annotation, which we strive to avoid in our work.
Second, among the self-supervised feature extraction functions, the \texttt{MSN}- and \texttt{DINO}-pretrained ViT backbones have the best trade-off in terms of both FID and IS. They even improve over the label-supervised backbones. %
This implies that the benefits of guidance is not unique to human annotated labels and self-supervised learning can provide a much more scalable alternative.
Since \texttt{DINO} \texttt{ViT-B/16} achieves the best FID performance, from now on we pick it as our self-supervised feature extraction function $g$.

\begin{figure*}
    \centering
    \includegraphics[width=0.96\textwidth]{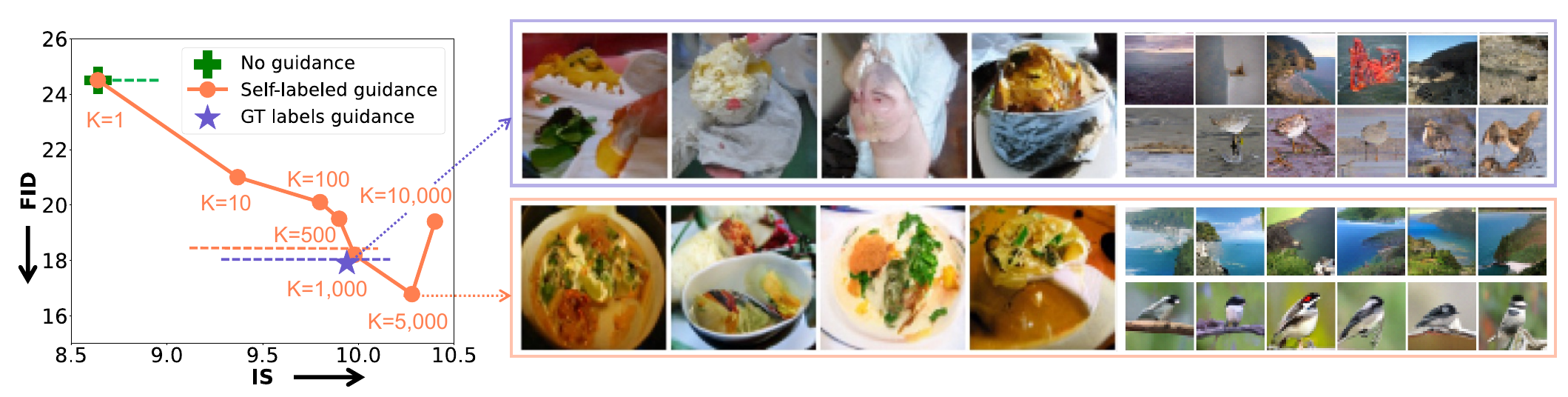}
    \caption{\textbf{Effect of number of clusters.} Self-labeled guidance outperforms DDPM without any guidance beyond a single cluster, is competitive with classifier-free guidance beyond 1{,}000 clusters and is even able to outperform guidance by ground-truth (GT) labels for 5{,}000 clusters. We visualize generated samples from ImageNet64 (middle) and ImageNet32 (right) for ground-truth labels guidance (top) and self-labeled guidance (bottom). More qualitative results  in~\Cref{sec:more_qualitative}
    .}
    \label{fig:ablate_clusterk} 
\end{figure*}

\paragraph{Effect of number of clusters.} 
Next, we ablate the influence of the number of clusters on the overall sampling quality. 
We consider 1 to 10{,}000 clusters on the extracted \texttt{CLS} token from the \texttt{DINO} \texttt{ViT-B/16} feature. For efficient comparison, we train each version for 20 epochs on ImageNet32. 
To put our sampling results in perspective, we also provide results for the no guidance and ground-truth guidance.

In~\Cref{fig:ablate_clusterk} we see that our model's performance improves monotonically as the cluster number increases from 1 to 5{,}000, consistently outperforming the no guidance baseline.
At 1{,}000 clusters, self-labeled guidance is competitive with the baseline trained using ground-truth labels. For 5{,}000 clusters, we find a sweet spot where our method outperforms the model using ground-truth labels, with an FID of 16.4 versus 17.9 and an IS of 10.35 versus 9.94. 
We can understand this result by considering (pseudo-)label conditioning as a method of transforming a single diffusion model into multiple specialized models, with each one focused on a distinct set of semantically coherent images.
Increasing the granularity of the groups, such as by increasing the number of clusters to 5{,}000, improves the semantic coherence of each group and simplifies the distribution. 
However, if the cluster number becomes too large, the self-supervised clusters may pick up on dataset-specific details that no longer correspond to general semantic concepts, leading to a deterioration in cluster quality and FID performance.
Nevertheless, we observe that samples generated from the same cluster ID exhibit high semantic coherence, indicating that the self-supervised clusters represent meaningful concepts that can be used to control the generation process. We discuss assigning semantic descriptions to clusters further in~\Cref{sec:more_qualitative}.

\begin{figure}%
    \centering
    \includegraphics[width=0.43\textwidth]{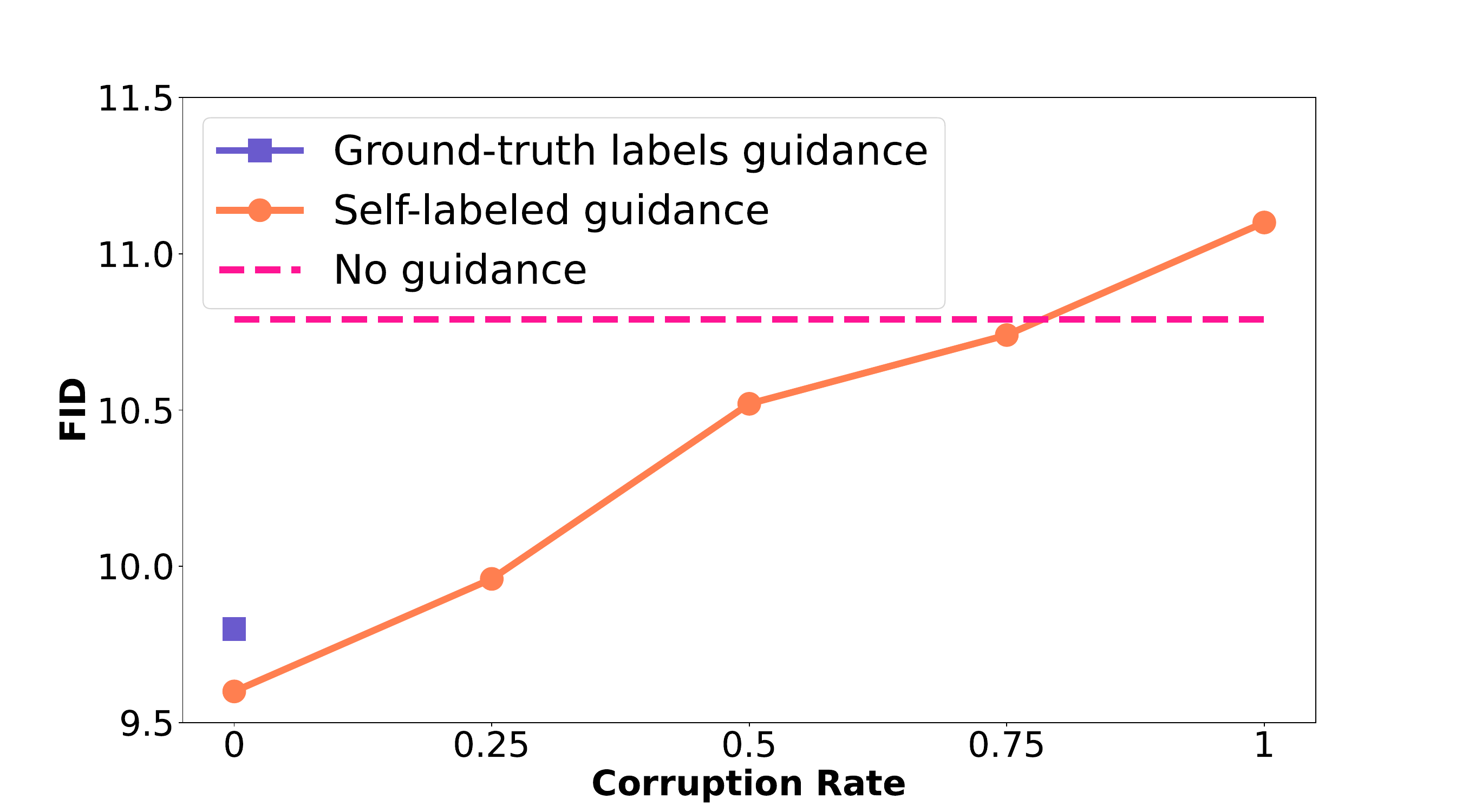}
    \caption{\textbf{Corruption of cluster assignments on CIFAR100.}
    Self-labeled guidance with 400 clusters outperforms the baseline trained with ground-truth labels. The FID performance deteriorates monotonically with the percentage of corrupted cluster assignments, underscoring the importance of assigning cluster ids via the self-supervised features.  
    }
    \label{fig:cifar100_corruption}
\end{figure}

\paragraph{Importance of self-supervised clusters.}
In the previous paragraph, we observed that training a diffusion model with 5{,}000 clusters can outperform the 1{,}000 ground-truth labels. Here, we check whether we can reproduce this result on another dataset and examine how the performance varies when we inject different degrees of noise into the cluster assignment. 
On CIFAR100~\citep{krizhevsky2009learning} we compare the ground-truth 100 labels with 400 self-supervised clusters. We corrupt the cluster assignments at different levels by randomly shuffling the cluster id for 25\% to 100\% of the images before training.
The results in~\Cref{fig:cifar100_corruption} highlight the importance of using self-supervised features for assigning clusters and that assigning cluster ids for a subset of the dataset is already sufficient to see improvements.

\begin{figure}
    \centering
    \includegraphics[width=0.45\textwidth]{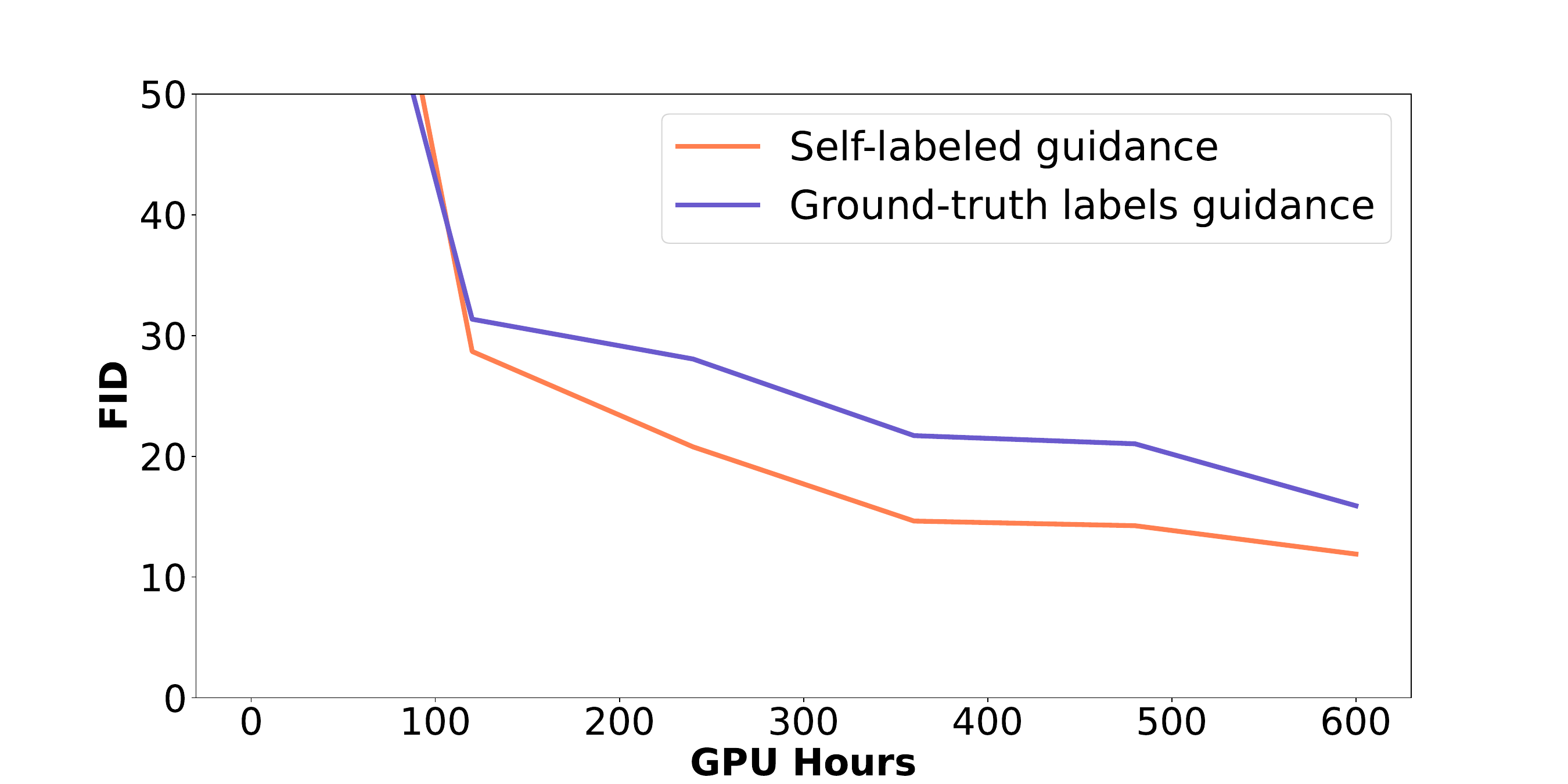}
    \caption{
    \textbf{Performance at different compute budgets.}
    After training for ${\sim}100$ GPU hours self-labeled guidance achieves persistent FID reduction over training with ground-truth labels.}
    \label{fig:gpuhour_vs_fid}
\end{figure}

\paragraph{Self-labeled comparisons on ImageNet32/64.} We compare our self-labeled guidance method against the baseline trained with ground-truth labels.
\textit{For a fair comparison, we use the same compute budget for all runs}. In particular, each model is trained for 100 epochs taking around 6 days on four RTX A5000 GPUs. %
Results on ImageNet32 and ImageNet64 are in~\Cref{tab:in32_in64}. 
Similar to \citet{dhariwal2021diffusion_beat}, we observe that any guidance setting improves considerably over the unconditional \& no-guidance model. Surprisingly, our self-labeled model even outperforms the ground-truth labels by a large gap in terms of FID of 1.9 and 4.7 points respectively. We hypothesize that the ground-truth taxonomy might be suboptimal for learning generative models and the self-supervised clusters offer a better guidance signal due to better alignment with the visual similarity of the images.
In~\Cref{fig:gpuhour_vs_fid} we report the FID at different training stages. It is worth
noting that the performance advantage of our self-guided method remains consistent over the entire training process.
The results suggest that the label-conditioned guidance from \cite{ho2021classifier} can be completely replaced by guidance from self-supervision, which would enable guided diffusion models to learn from even larger (unlabeled) datasets than feasible today. 

\begin{table}
\centering

\resizebox{1.0\columnwidth}{!}{%
\begin{tabular}{l@{\hspace{0pt}}c rrrr}
\toprule
&\multirow{2}{*}{\makecell{\textbf{Annotation}\\ \textbf{free?}}}& \multicolumn{2}{c}{\textbf{ImageNet32}} & \multicolumn{2}{c}{\textbf{ImageNet64}}  \\

\cmidrule(lr){3-4} \cmidrule(lr){5-6}
 \textbf{Diffusion Method} && FID$\downarrow$ & IS $\uparrow$ & FID$\downarrow$ & IS$\uparrow$   \\
 \midrule
 \color{gray} Ground-truth labels guidance & \color{gray}\xmark & \color{gray}\ttt{$9.2$}& \color{gray}$19.0$ & \color{gray}\ttt{$16.8$} & \color{gray}$18.6$   \\
No guidance & \cmark & $14.3$ &  $10.8$  & \ttt{${36.1}$}  & $10.4$ \\
 Self-labeled guidance & \cmark & $\textbf{7.3}$ & $\textbf{20.3}$  & $\textbf{12.1}$ & $\textbf{23.1}$   \\ 
 \bottomrule
\end{tabular}
}

\caption{\textbf{Self-labeled comparisons on ImageNet32/64.} Self-labeled guidance surpasses the no-guidance baseline by a large margin on both datasets and even outperforms the guided diffusion model trained using ground-truth class labels.
}
\label{tab:in32_in64}
\end{table}

\paragraph{Higher resolution image generation.}
Finally, we verify the effectiveness of self-labeled guidance on larger images. We report on the ImageNet-100~\cite{tian2020contrastive} (a subset of ImageNet-1k with 100 classes) and the LSUN-Churches dataset~\cite{yu15lsun}, both with images of size $256 \times 256$. Notably, the latter does not come with any annotations rendering a ground-truth guided baseline infeasible.
Too limit computation, we use the Latent Diffusion Model~\cite{rombach2022high_latentdiffusion_ldm} which is much more efficient at training with large image sizes than directly learning a diffusion model in the pixel space.

\begin{figure}
    \centering
    \includegraphics[width=0.44\textwidth]{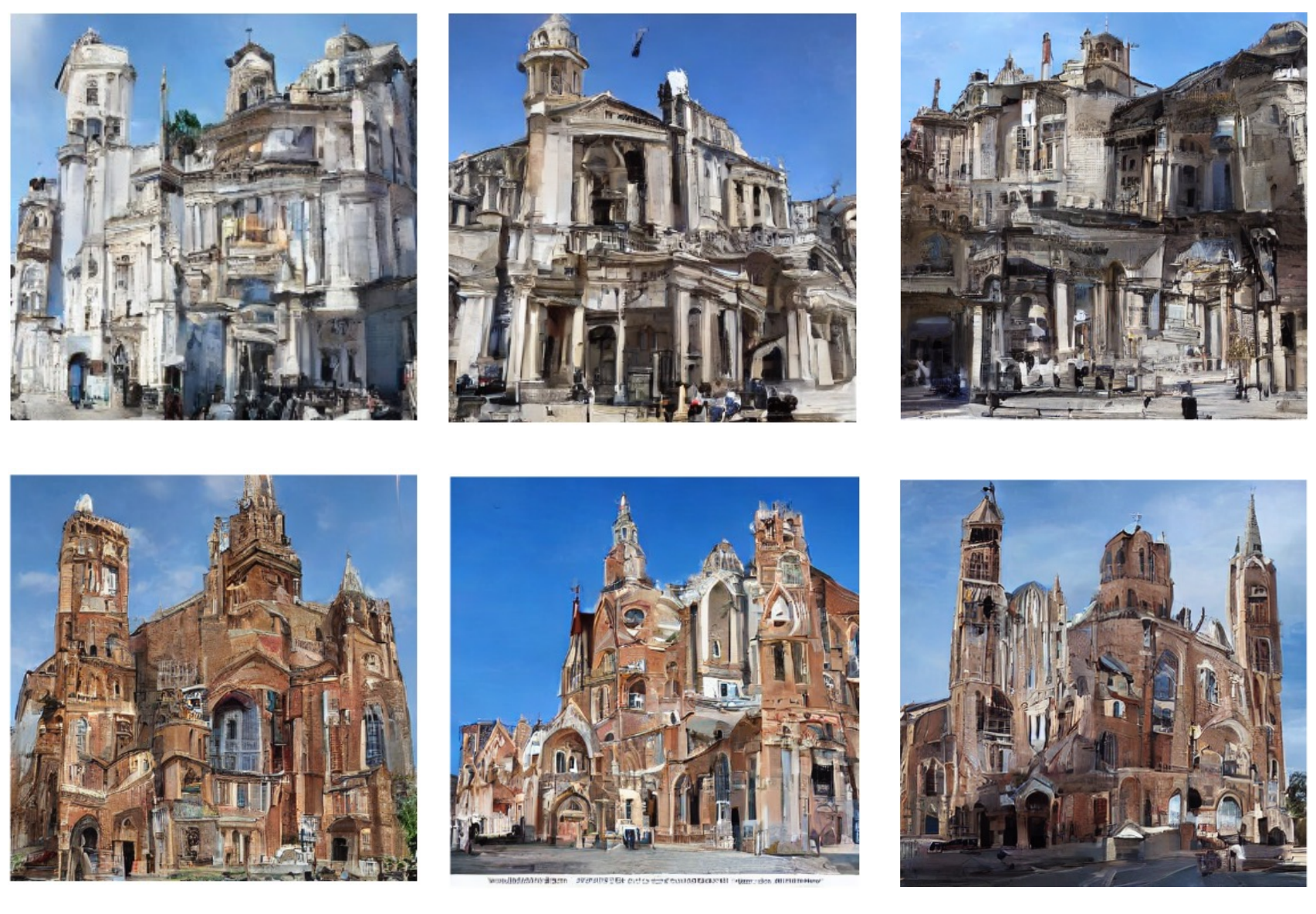}
    \vspace{-4mm} 
    \caption{\textbf{Generated samples at $256 \times 256$ resolution using self-labeled guidance on LSUN-Churches.} 
    Samples in each row are from the same cluster. Self-labeled guidance enables semantically coherent samples, despite the absence of ground-truth annotations.}
    \label{fig:churches256_vis}
\end{figure}

\Cref{tab:longer training high resolution} shows that self-labeled guidance significantly outperforms the baselines, indicating the effectiveness of our method for high-resolution images. Note that the lack of ground-truth labels for LSUN-Churches reflects an advantage of our method since most real-world images are unlabeled.
We show generated samples for two different clusters in~\Cref{fig:churches256_vis}. The samples are diverse and reflect shared characteristics for samples guided by the same cluster. For more qualitative and quantitative results we refer to~\Cref{sec:more_result_supp}.

\begin{table}
\centering
\resizebox{1.0\columnwidth}{!}{%
\begin{tabular}{l@{\hspace{0pt}}c rrc}
\toprule
&\multirow{2}{*}{\makecell{\textbf{Annotation}\\ \textbf{free?}}}& \multicolumn{2}{c}{\textbf{ImageNet-100}} & \multicolumn{1}{c}{\textbf{LSUN-Churches}}  \\

\cmidrule(lr){3-4} \cmidrule(lr){5-5}
 \textbf{Diffusion Method} && FID$\downarrow$ & IS $\uparrow$ & FID$\downarrow$  \\
 \midrule
 \color{gray} Ground-truth labels guidance & \color{gray}\xmark & \color{gray}\ttt{$21.2$}& \color{gray}$64.1$ & \color{gray}---  \\
No guidance & \cmark & $42.1$ &  $41.1$  & $19.2$ \\
 Self-labeled guidance & \cmark & $\textbf{16.1}$ & $\textbf{78.3}$  & $\textbf{15.2}$  \\ 
 \bottomrule
\end{tabular}
}

\caption{\textbf{Higher resolution image generation.} ImageNet-100 and LSUN-Churches results for images of size 256$\times$256.}
\label{tab:longer training high resolution}
\end{table}

\subsection{Self-Boxed Guidance}

We run experiments on Pascal VOC and COCO\_20K to validate the efficacy of self-boxed guidance. The self-boxed guidance model takes a bounding box in addition to the cluster-ID as guidance signal. To obtain the class-agnostic object bounding boxes, we use LOST~\citep{simeoni2021localizing_lost} as our self-annotation function $f$ in~\Cref{eq:self_guidance}. For the clustering, we empirically found $k{=}100$ to work well for both datasets, as both are relatively small in scale when compared to ImageNet. 
We train our diffusion model for 800 epochs with images 
 of size 64\texttimes 64.
We report train FID for Pascal VOC and train/validation FID for COCO\_20K. We evaluate the performance on the validation split by extracting the guidance signal from the training dataset to ensure that there is no information leakage.
See~\Cref{sec:hyperparames} for more details.

\begin{figure*}
    \centering
    \includegraphics[width=0.87\textwidth]{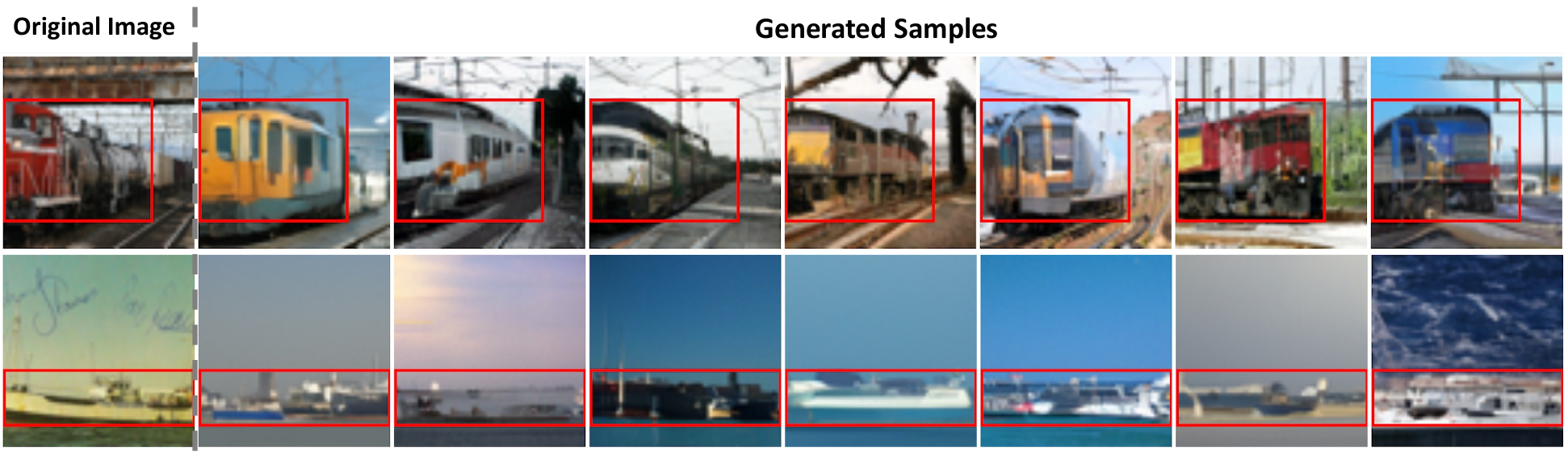}
    \caption{\textbf{Self-boxed guided diffusion results on Pascal VOC.} Each column is sampled using different random noise. Our method generates visually diverse and semantically consistent images. The image-level guidance signal successfully puts a limit on the model to create \textit{train station, port} scenarios. The backgrounds are realistic and in harmony with the guidance boxes.
    }
    \label{fig:voc_lost_guidance}
\end{figure*}

\paragraph{Self-boxed comparisons on Pascal VOC and COCO\_20K.} 
For the ground-truth labels guidance baseline, we condition on a class embedding. Since there are now multiple objects per image, we represent the ground-truth class with a multi-hot embedding.
Aside from the class embedding which is multi-hot in our method, all other settings remain the same for a fair comparison. 
The results in~\Cref{tab:lost_guidance}, confirm that the multi-hot class embedding is indeed effective for multi-label datasets, improving over the no-guidance model by a large margin. This improvement comes at the cost of manually annotating  multiple classes per image. Self-boxed guidance further improves upon this result, by reducing the FID by an additional $5.1$ and $3.3$ points respectively without using any ground-truth annotation. In~\Cref{fig:voc_lost_guidance}, we show our method generates diverse and semantically well-aligned images.

\begin{table}
\centering

\resizebox{\columnwidth}{!}{%
\setlength\tabcolsep{3pt}
\begin{tabular}{l@{\hspace{0pt}}cccc}
\toprule
&\multirow{2}{*}{\makecell{\textbf{Annotation}\\ \textbf{free?}}}&\multirow{2}{*}{\textbf{Pascal VOC}} & \multirow{2}{*}{\textbf{COCO\_20K}}   \\ 
\textbf{Diffusion Method} &&& \\
\midrule
\color{gray} Ground-truth labels guidance &\xmark &  \color{gray} $23.5$ &  \color{gray}$19.3$ \\ 
  No guidance & \cmark & $58.6$  &  $42.5$\\
    Self-boxed guidance &\cmark & $\textbf{18.4}$ & $\textbf{16.0}$\\
    \color{gray} Ground-truth boxes guidance &\xmark &  \color{gray} $13.2$ &  \color{gray} $9.6$ \\ 
 \bottomrule
\end{tabular}
}
\caption{\textbf{Self-boxed comparisons on Pascal VOC and COCO\_20K.} Self-boxed guidance outperforms the no-guidance baseline FID considerably for multi-label datasets and is even better than a label-supervised alternative.
}

\label{tab:lost_guidance}
\end{table}

\begin{table}
\centering

\resizebox{\columnwidth}{!}{%
\setlength\tabcolsep{3pt}
\begin{tabular}{l@{\hspace{0pt}}ccccc}
\toprule
&\multirow{2}{*}{\makecell{\textbf{Annotation}\\ \textbf{free?}}}& \textbf{Pascal VOC} & \multicolumn{2}{c}{\textbf{COCO-Stuff}} \\
\cmidrule(lr){4-5}
\textbf{Diffusion Method} &&& Train & Val \\
 \midrule
  \color{gray}Ground-truth labels guidance&\xmark &   \color{gray}$23.5$ & \color{gray}$16.3$  & \color{gray}$20.5$ \\ 
  No guidance& \cmark  &$58.6$ & $29.1$& $34.1$ \\
Self-segmented guidance & \cmark & $\textbf{17.1}$  & $\textbf{12.5}$& $\textbf{17.7}$ \\ 
\color{gray}Ground-truth masks & \xmark & \color{gray}$12.5$ & \color{gray}$\ttt{8.1}$ &\color{gray}$\ttt{11.2}$ \\ 
 \bottomrule
\end{tabular}
}

\caption{\textbf{Self-segmented comparisons on Pascal VOC and COCO-Stuff.}
Any form of guidance results in a considerable FID reduction over the no-guidance model. Self-segmented guidance improves over ground-truth multi-hot labels guidance and narrows the gap with guidance by annotation-intensive ground-truth masks.
}
\vspace{-2mm}
\label{tab:stego_guidance}
\end{table}

\begin{figure*}
    \centering
    \includegraphics[width=0.93\textwidth]{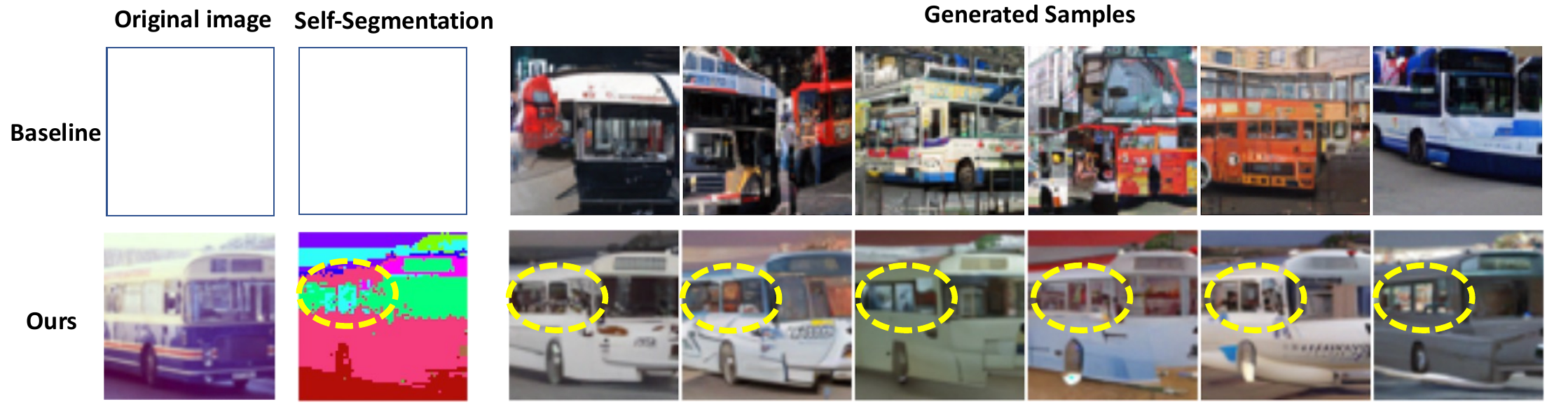}
    \caption{\textbf{Self-segmented guided diffusion results on Pascal VOC.} Each column is sampled using different random noise.
    The  visualization indicates our self-segmented guidance provides more fine-grained guidance than ground-truth labels guidance for the generation of bus images. Note how the noisy window-bar in the self-segmented mask (marked by dotted ellipse) still results in plausible window separations in the generated image samples.
    We provide more examples in~\Cref{sec:more_qualitative}
    . }
    \label{fig:voc_steo_vis}
\end{figure*}

\subsection{Self-Segmented Guidance}
Finally, we validate the efficacy of self-segmented guidance on Pascal VOC and COCO-Stuff. For COCO-Stuff we follow the split from~\citep{hamilton2022unsupervised_stego, ji2019invariant_iic,cho2021picie,zhang2022dense_dsn}, with a train set of 49,629 images and a validation set of 2,175 images. Classes are merged into 27 (15 stuff and 12 things) categories. For self-segmented guidance, we apply \texttt{STEGO}~\citep{hamilton2022unsupervised_stego} as our self-annotation function $f$ in~\Cref{eq:self_guidance}.  
We set the cluster number to 27 for COCO-Stuff, and 21 for Pascal VOC, following~\texttt{STEGO}. We train all models on images of size 64\texttimes 64, for 800 epochs on Pascal VOC, and for 400 epochs on COCO-Stuff. 
We report the train FID for Pascal VOC and both train and validation FID for COCO-Stuff. %
More details on the dataset and experimental setup are provided in~\Cref{sec:hyperparames}.

\vspace{-1mm}
\paragraph{Self-segmented comparisons on Pascal VOC and COCO-Stuff.}  
We compare against both the ground-truth labels guidance baseline from the previous section and a model trained with ground-truth semantic masks guidance. The results in~\Cref{tab:stego_guidance} demonstrate that our self-segmented guidance still outperforms the ground-truth labels guidance baseline on both datasets. 
The comparison between ground-truth labels and segmentation masks reveals an improvement in image quality when using the more fine-grained segmentation mask as the condition signal.
But these segmentation masks are one of the most costly types of image annotations that require every pixel to be labeled. Our self-segmented approach avoids the necessity for annotations while narrowing the performance gap, and more importantly offering fine-grained control over the image layout. We demonstrate this controllability with examples in~\Cref{fig:voc_steo_vis} and explain how to assign semantic descriptions to the clusters in~\Cref{sec:more_qualitative}. These examples further highlight a robustness against noise in the segmentation masks, which our method acquires naturally due to training with noisy segmentations.

\section{Conclusion}

We have explored the potential of self-supervision signals for diffusion models and propose a framework for self-guided diffusion models. By leveraging a feature extraction function and a self-annotation function, our framework provides guidance signals at various image granularities: from the level of holistic images to object boxes and even segmentation masks. Our experiments indicate that self-supervision signals are an adequate replacement for existing guidance methods that generate images by relying on annotated image-label pairs during training. Furthermore, both self-boxed and self-segmented approaches demonstrate that we can acquire fine-grained control over the image content, without any ground-truth bounding boxes or segmentation masks. 
Though in certain cases, clusters can capture visual concepts that are challenging to articulate in everyday language, such as in the case of LSUN-Churches.
For future research, it would be interesting to investigate the efficacy of our self-guidance approach on feature extractors trained on larger datasets or with image-text pairs~\cite{radford2021learning_clip}.
Ultimately, our goal is to enable the benefits of self-guided diffusion for unlabeled and more diverse datasets at scale, wherein we believe this work is a promising first step.

{\small
\paragraph{Acknowledgements.}
The work of DWZ is part of the research programme Perspectief EDL with project number P16-25 project 3, which is financed by the Dutch Research Council (NWO) domain Applied and Engineering Sciences (TTW). We thank EscherCloud AI for the European compute resources.
}

\newpage

{\small
\bibliographystyle{ieee_fullname}
\bibliography{egbib}
}
\appendix

\clearpage

\tableofcontents

\clearpage

\section{More results and details}
\label{sec:more_result_supp}

\subsection{Results on Churches-256}
\label{sec:churches_result_supp}

\paragraph{Implementation details.} We directly use the official code of Latent Diffusion Model\footnote{https://github.com/CompVis/latent-diffusion}~\citep{rombach2022high_latentdiffusion_ldm}, and reduce the base channel number from 192 to 128 and attention resolution from [32, 16, 8, 4] to [8, 4] to accelerate training. Note that these changes significantly reduce the number of parameters from 294M to 108M.

\begin{figure*}
    \centering
    \includegraphics[width=0.97\textwidth]{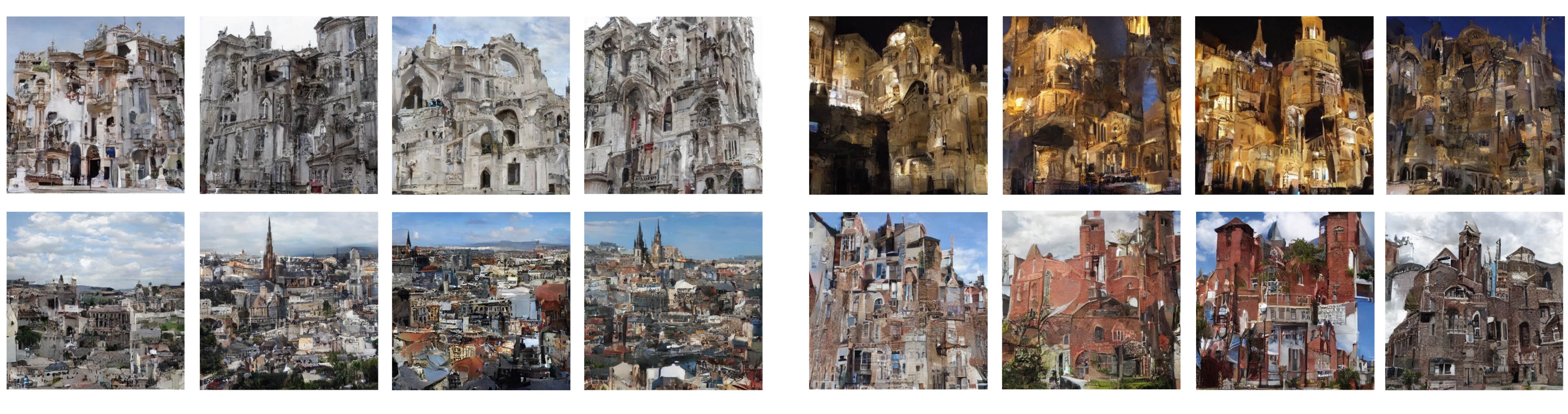}
    \caption{\textbf{Generated samples using self-labeled guidance on LSUN-Churches 256$\times$256.} Each row corresponds to a different cluster. Clusters can capture concepts like nighttime, a far shot that includes the city, a close shot of the church, and the church's color.}
    \label{fig:churches_cluster_vis}
\end{figure*}

\paragraph{Qualitative and Quantitative result.} We present more qualitative results in~\Cref{fig:churches_cluster_vis}. We use the FID metric for quantitative comparison. For non-guidance and our self-labeled guidance, we get an FID of 23.1 and 16.2 respectively. Our self-labeled guidance improves by almost 7 points for free.

\subsection{Precision and Recall in ImageNet32/64 dataset}

We show the extra  results of ImageNet on precision and recall in~\Cref{tab:in32_in64_full_result}. We follow the evaluation code of precision and recall from ICGAN~\citep{icgan}, our self-labeled guidance also outperforms ground-truth labels in precision and remains competitive in the recall. %

\begin{table*}
\centering
\begin{tabular}{lcrrrrrrrr}
\toprule
\textbf{Diffusion Method}&\textbf{Annotation-free?} & \multicolumn{4}{c}{\textbf{ImageNet32}} & \multicolumn{4}{c}{\textbf{ImageNet64}}  \\

\cmidrule(lr){3-6} \cmidrule(lr){7-10}
 && FID$\downarrow$ &IS $\uparrow$ &P $\uparrow$ & R $\uparrow$ & FID$\downarrow$ & IS$\uparrow$ &P$\uparrow$ & R $\uparrow$ \\
 \midrule
\color{gray}Ground-truth labels guidance & \xmark & \color{gray}\ttt{$9.2$} & \color{gray}$19.0$ &\color{gray}$0.71$ & \color{gray}$0.62$ & \color{gray}\ttt{$16.8$} & \color{gray}$18.6$ &\color{gray}$0.71$ & \color{gray}$\textbf{0.62}$   \\
No guidance &\cmark & $14.3$ & $10.8$   & $0.49$& $0.61$  & $36.1$  & $10.4$ &$0.59$& $0.60$ \\
 Self-labeled guidance  & \cmark & $\textbf{7.3}$ & $\textbf{20.3}$  &  $\textbf{0.77}$& $\textbf{0.63}$& $\textbf{12.1}$ & $\textbf{23.1}$  &$\textbf{0.78}$& $\textbf{0.62}$   \\ 
 \bottomrule
\end{tabular}
\caption{\textbf{Comparison with baseline on ImageNet32 and ImageNet64 dataset with FID, IS, Precision (P), Recall (R).}}
\label{tab:in32_in64_full_result}
\end{table*}

\begin{figure}%
    \centering
    \includegraphics[width=0.4\textwidth]{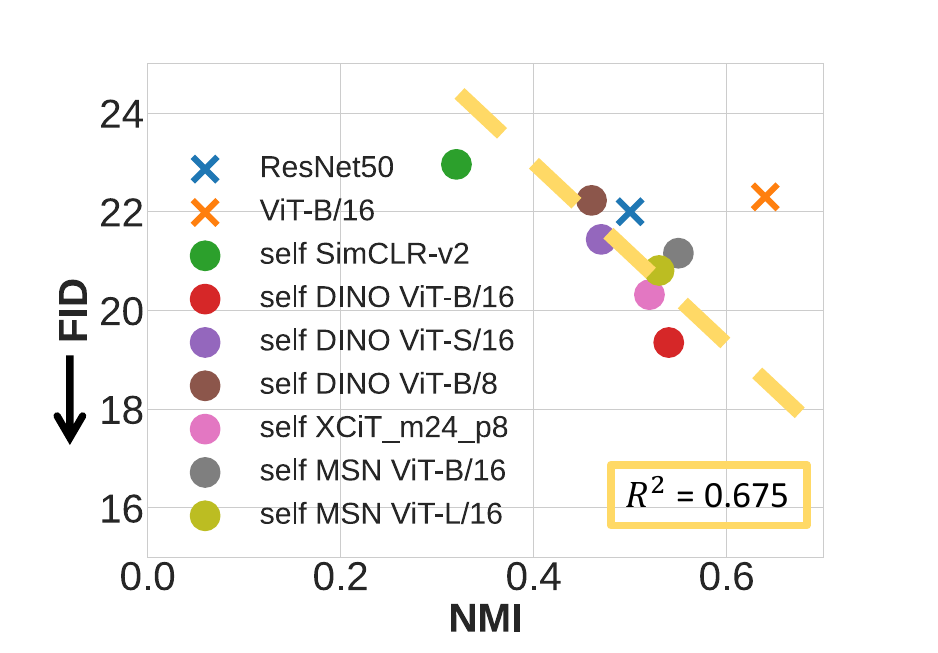}
    \vspace{-2mm}
    \caption{\textbf{Correlation between NMI and FID} on ImageNet32. The Normalized Mutual Information (NMI) is not related to FID for supervised backbones,  while, for the self-supervised model, NMI and FID are negatively correlated.
	}
    \label{tab:nmi_fid_correalation}
\end{figure}

\subsection{Correlation between NMI and FID in different feature backbones.} 
Normalized Mutual Information (NMI) can be used to assess the performance in self-supervised representation learning. It measures the similarity between the cluster assignments and the ground-truth labels. We examine whether there is a relation between the quality of the self-supervised method, as it is typically measured, and the FID resulting from the clusters induced by the self-supervised features.
In~\Cref{tab:nmi_fid_correalation} we plot the NMI and FID for different self-supervised models. The models trained with ground-truth labels show no change in FID for different NMI values. In contrast, the self-supervised models exhibit a negative correlation between the NMI and FID, suggesting that NMI is also predictive of the model's usefulness in our setting.
This indicates that future progress in self-supervised learning will also translate to improvements to self-labeled guidance.

\subsection{Varying guidance strength $w$}

\begin{figure*}
\centering
\begin{subfigure}{.45\textwidth}
    \centering
    \includegraphics[width=0.8\textwidth]{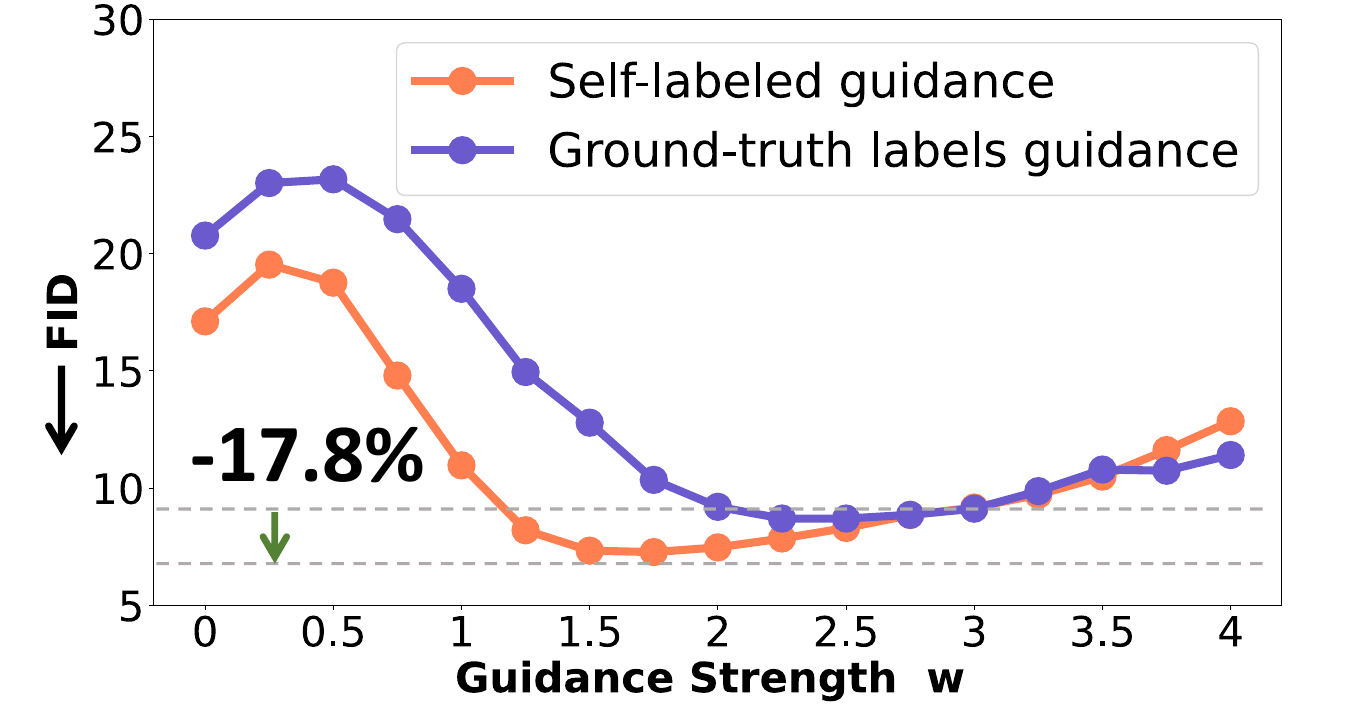}
    \caption{Setting I: ImageNet32 balanced }
    \label{fig:cond_scale_in32}
\end{subfigure}%
\begin{subfigure}{.45\textwidth}
    \centering
    \includegraphics[width=0.8\textwidth]{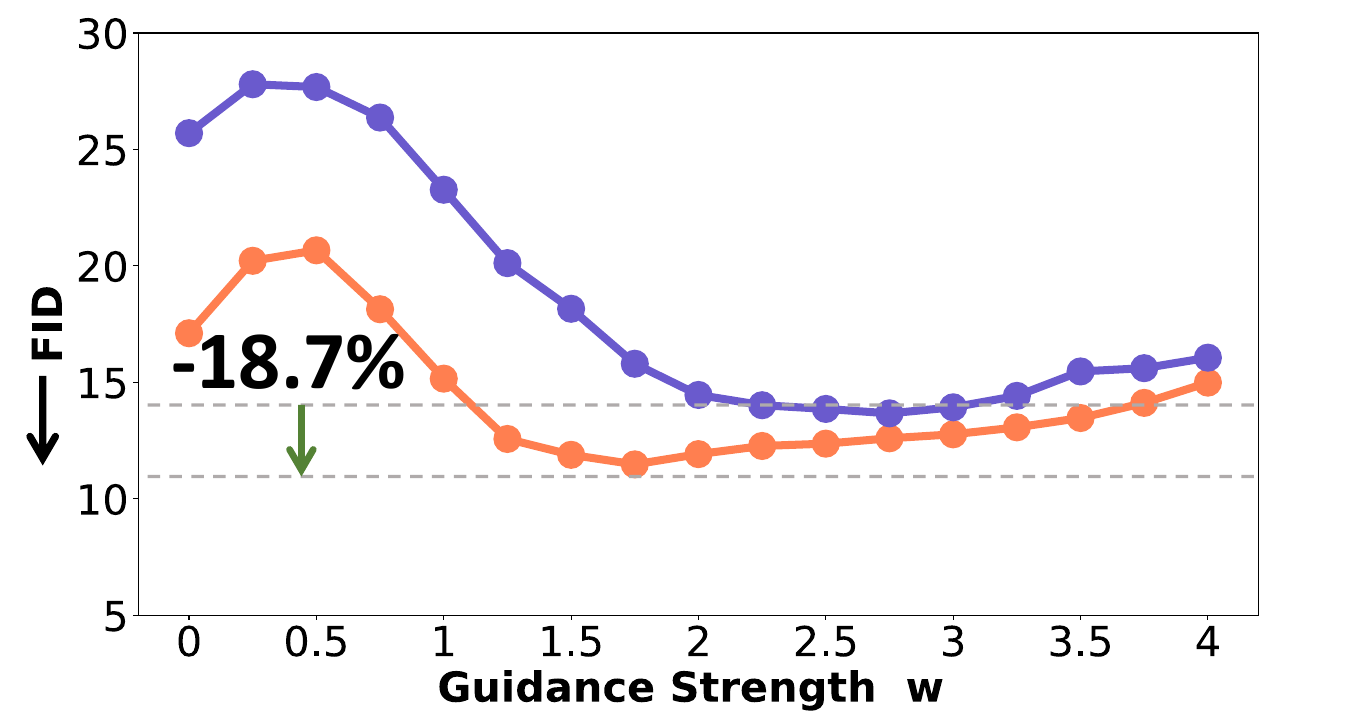}
    \caption{Setting II: ImageNet32 unbalanced}
    \label{fig:cond_scale_in32_lt}
\end{subfigure}
\caption{\textbf{Varying guidance strength ${w}$.} Self-labeled guidance surpasses the guidance based on ground-truth labels for both (a) ImageNet32 balanced and (b) ImageNet32 unbalanced.
The dotted gray line indicates the best-achieved performance of both methods under various guidance strengths.
The difference between them is slightly more prominent for unbalanced data, 
we conjecture that this is because our self-labeled guidance is obtained by clustering based on the statistics of the overall dataset, which can potentially lead to more robust performance in unbalanced setting.}
\label{fig:FID}
\end{figure*}

We consider the influence of the guidance strength $w$ on our sampling results.
We mainly conduct this experiment in ImageNet32,
as the validation set of ImageNet32 is strictly balanced, 
we also consider an unbalanced setting which is more similar to real-world deployment. Under both settings, we compare the FID between our self-labeled guidance and ground-truth guidance.  We train both models for 100 epochs.
For the standard ImageNet32 validation setting in~\Cref{fig:cond_scale_in32}, our method achieves a 17.8\% improvement for the respective optimal guidance strength of the two methods. Self-labeled guidance is especially effective for lower values of $w$. 
We observe similar trends for the unbalanced setting in ~\Cref{fig:cond_scale_in32_lt}, be it that the overall FID results are slightly higher for both methods. The improvement increases to 18.7\%. 
We conjecture this is due to the unbalanced nature of the $k$-means  algorithm~\citep{last2017oversampling}, and clustering based on the statistics of the overall dataset can potentially lead to more robust performance in an unbalanced setting.

\subsection{Cluster number ablation in self-boxed \\guidance}

 In~\cref{tab:voc64_box_ablate_clusterk}, we empirically evaluate the performance when we alter the cluster number in our self-boxed guidance. We find the performance will increase from $k=21$ to $k=100$, and saturated at $k=100$.

\begin{table}[h]
\centering
\begin{tabular}{c|c}
\toprule
 \textbf{Cluster number} $k$  & \textbf{FID} $\downarrow$ \\
 \midrule
  $21$&  $22.5$ \\
  $50$&  $18.6$ \\
  $100$&  $18.5$ \\
 \bottomrule
\end{tabular}
\caption{\textbf{Cluster number ablation on Pascal VOC dataset for self-boxed guidance.}} 
\label{tab:voc64_box_ablate_clusterk}
\end{table}

\subsection{Trend visualization of training loss  and \\validation FID}

\begin{figure}
    \centering
    \includegraphics[width=0.47\textwidth]{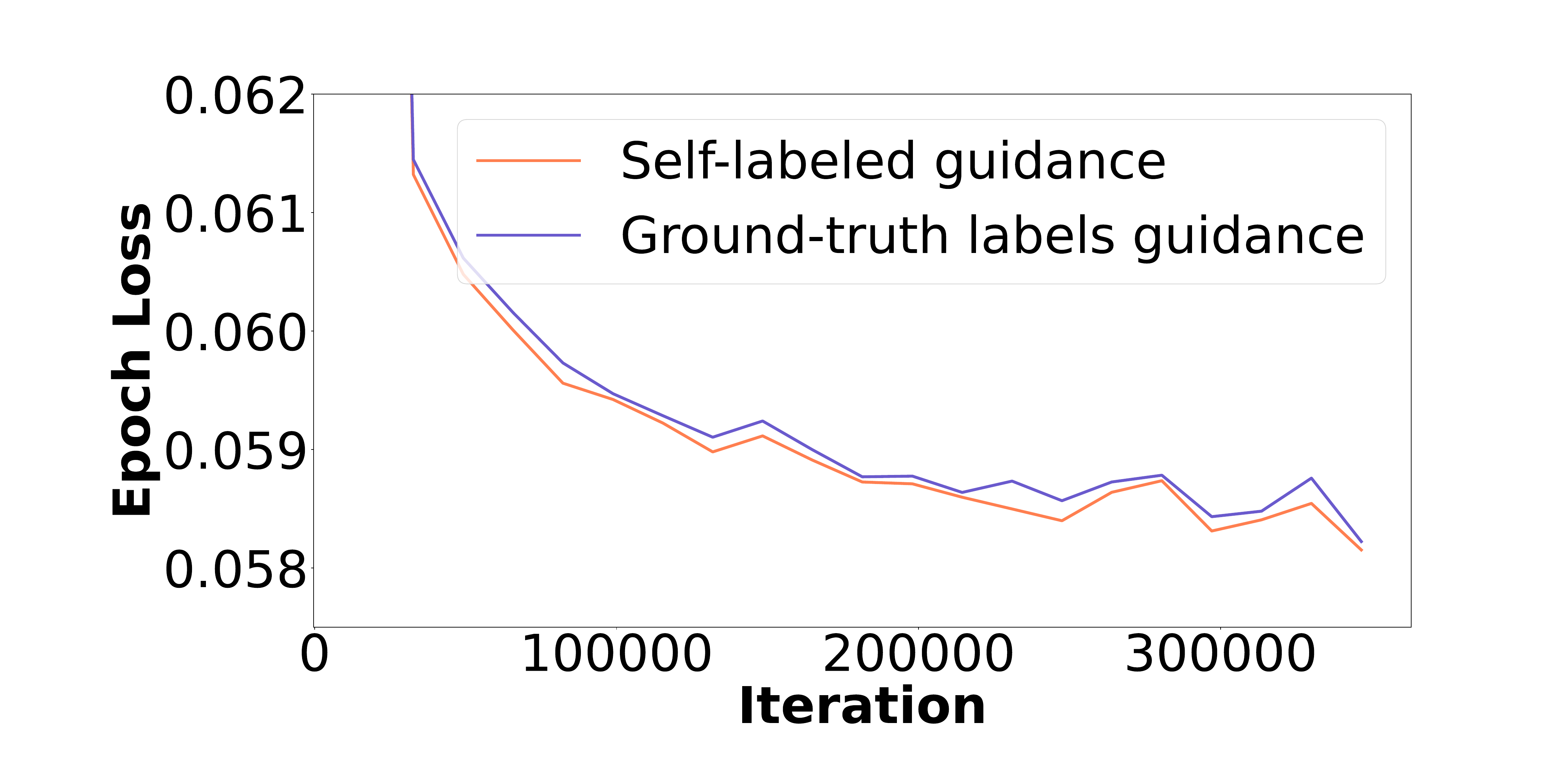}
    \caption{\textbf{Epoch loss trend.}}
    \label{fig:trend_loss_fid}
\end{figure}

We visualize the trend of training loss and validation FID in~\Cref{fig:trend_loss_fid}.

\section{More experimental details}
\label{sec:hyperparames}

\textbf{Training details.} For our best results,  we train 100 epochs on 4 GPUs of A5000 (24G) in ImageNet. We train 800/800/400 epochs on 1GPU of A6000 (48G) in Pascal VOC, COCO\_20K, and COCO-Stuff, respectively. All qualitative results in this paper are trained in the same setting as mentioned above. We train and evaluate the Pascal VOC, COCO\_20K, and COCO-Stuff  in image size 64, and visualize them by bilinear upsampling to 256, following ~\citep{liu2022compositional_diff}.

\textbf{Sampling details.} We sample the guidance signal from the distribution of training set in our all experiments.  For each timestep, we need twice of Number of Forward Evaluation (NFE), we optimize them by concatenating the conditional and unconditional signal along the batch dimension so that we only need one time of NFE in every timestep.

\textbf{Evaluation details.} 
We use the common package Clean-FID~\citep{parmar2021cleanfid}, torch-fidelity~\citep{obukhov2020torchfidelity} for FID, IS calculation, respectively.
For IS, we use the standard 10-split setting,
we  only report IS on ImageNet, as it might be not an appropriate metric for non object-centric datasets~\citep{barratt2018note}.
 For the checkpoint, we pick the checking point every 10 epochs by minimal FID between generated sample set and the train set.

\begin{figure*}
    \centering
    \includegraphics[width=0.99\textwidth]{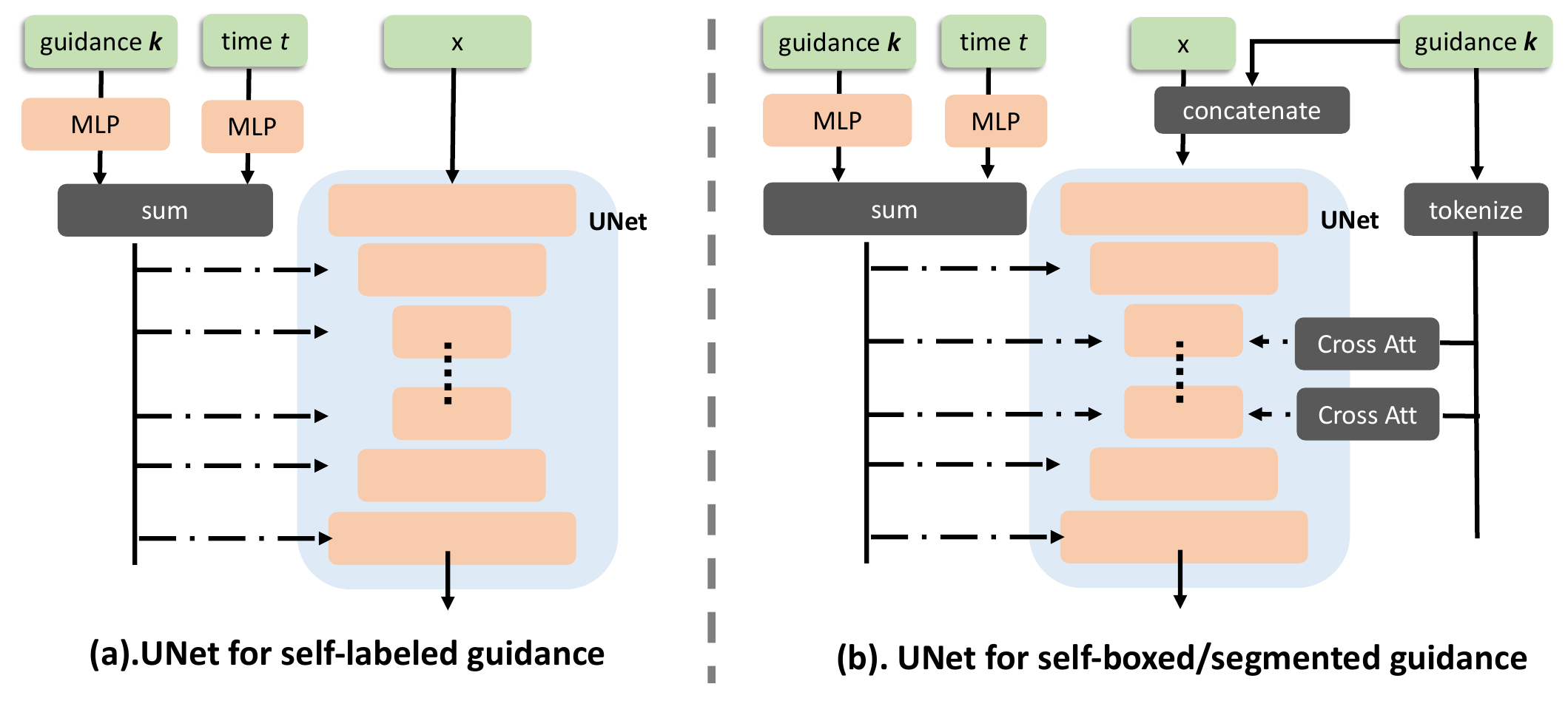}
    \caption{\textbf{The structure of UNet module.} }
    \label{fig:unet_module}
\end{figure*}

\subsection{UNet structure}
\label{sec:unet_structure_details}

\paragraph{Guidance signal injection.} We describe the detail of guidance signal injection in~\Cref{fig:unet_module}. The injection of self-labeled guidance and self-boxed/segmented guidance is slightly different. The common part is by concatenation between timestep embedding and noisy input, the concatenated feature will be sent to every block of the UNet.
 For the self-boxed/segmented guidance, we not only conduct the information fusion as above but also incorporate the spatial inductive-bias by concatenating it with input, the concatenated result will be fed into the UNet.

\paragraph{Timestep embedding.} We embed the raw timestep information by two-layer MLP: FC(512, 128)$\rightarrow $SiLU$\rightarrow $FC(128, 128).

\paragraph{Guidance embedding.} The guidance is in the form of one/multi-hot embedding $\mathbb{R}^K$, we feed it into two-layer MLP: FC(K, 256)$\rightarrow $SiLU$\rightarrow $FC(256, 256), then feed those guidance signal into the UNet  following in~\Cref{fig:unet_module}.

 \paragraph{Cross-attention.} In training for non object-centric dataset, we also tokenize the guidance signal to several tokens following  Imagen~\citet{saharia2022photorealistic_imagen}, we concatenate those tokens with image tokens (can be transposed to a token from typical feature map by $\mathbb{R}^{W\times H \times C} \to \mathbb{R}^{C \times WH}$), the cross-attention~\citep{rombach2022high_latentdiffusion_ldm,blattmann2022retrieval} is conducted by \texttt{CA}($\texttt{m}$, \texttt{concat}$[\bk,\texttt{m}]$). %
Due to the quadratic complexity of transformer~\citep{katharopoulos2020transformers_quadratic,lu2021soft_quadratic2}, we only apply the cross-attention in lower-resolution feature maps.

\subsection{Training Parameter}
\label{sec:unet_parameter_details}

\begin{table*}
\centering
\small
\begin{tabular}{ll}
\toprule
Base channels: 128                        & Optimizer: AdamW  \\
Channel multipliers: 1, 2, 4      & Learning rate: $3e-4$ \\
Blocks per resolution: 2                  & Batch size: 128 \\
Attention resolutions: 4          & EMA: 0.9999 \\
number of head: 8             & Dropout: 0.0 \\
Conditioning embedding dimension: 256    & Training hardware: 4 $\times$  A5000(24G) \\
Conditioning embedding MLP layers: 2      & Training Epochs: 100 \\
Diffusion noise schedule: linear          &  Weight decay: 0.01   \\
Sampling timesteps: 256 & \\
\bottomrule
\end{tabular}
\caption{\textbf{ 3$\times$32$\times$32 model, 4GPU, ImageNet32.}}
\end{table*}

\begin{table*}
\centering
\small
\begin{tabular}{ll}
\toprule
Base channels: 128                       & Optimizer: AdamW  \\
Channel multipliers: 1, 2,  4      & Learning rate: $1e-4$ \\
Blocks per resolution: 2                  & Batch size: 48 \\
Attention resolutions: 4        & EMA: 0.9999 \\
number of head: 8             & Dropout: 0.0 \\
Conditioning embedding dimension: 256    & Training hardware: 4 $\times$  A5000(24G) \\
Conditioning embedding MLP layers: 2      & Training Epochs: 100 \\
Diffusion noise schedule: linear          &  Weight decay: 0.01   \\
Sampling timesteps: 256 \\
\bottomrule
\end{tabular}
\caption{\textbf{ 3$\times$64$\times$64 model, 4GPU, ImageNet64.}}
\end{table*}

\begin{table*}
\small
\centering
\begin{tabular}{ll}
\toprule
Base channels: 128                       & Optimizer: AdamW  \\
Channel multipliers: 1, 2, 4      & Learning rate: $1e-4$ \\
Blocks per resolution: 2                  & Batch size: 80 \\
Attention resolutions: 4          & EMA: 0.9999 \\
Number of head: 8             & Dropout: 0.0 \\
Conditioning embedding dimension: 256    & Training hardware: 1 $\times$  A6000(45G) \\
Conditioning embedding MLP layers: 2      & Training Epochs: 800/800/400 \\
Diffusion noise schedule: linear          &  Weight decay: 0.01   \\
Sampling timesteps: 256 &  Context token number: 8\\
Context dim: 32 &\\
\bottomrule
\end{tabular}
\caption{\textbf{ 3$\times$64$\times$64 model, 1GPU, Pascal VOC, COCO\_20K, COCO-Stuff.}}
\label{tab:unet_param_detail}
\end{table*}

\begin{table*}
\small
\centering
\begin{tabular}{ll}
\toprule
Base channels: 128                       & Optimizer: AdamW  \\
Channel multipliers: 1, 2,2,3, 4      & Learning rate: $5e-5$ \\
Blocks per resolution: 2                  & Batch size: 48 \\
Attention resolutions: 4,8          & EMA: 0.9999 \\
Number of head: 8             & Dropout: 0.0 \\
Conditioning embedding dimension: 256    & Training hardware: 4 $\times$  A5000(24G) \\
Conditioning embedding MLP layers: 2      & Training Steps: 600k \\
Diffusion noise schedule: linear          &  Weight decay: 0.01   \\
Sampling timesteps: 200 &  \\
\bottomrule
\end{tabular}
\caption{\textbf{ 3$\times$256$\times$256 model, 4GPU, Churches-256.}}
\label{tab:unet_param_detail}
\end{table*}

\subsection{Dataset preparation}
\label{sec:dataset_preparation}

\paragraph{The preparation of unbalanced dataset.}\label{sec:imbalance_dataset_prep} There are 50,000 images in the validation set of ImageNet with 1,000 classes (50 instances for each). We index the class from 0 to 999, for each class $c_{i}$, the instance of the class $c_{i}$ is $\lfloor i \times50/1000 \rfloor =\lfloor i/200 \rfloor $.

\paragraph{Pascal VOC.}

We use the standard split from~\citep{simeoni2021localizing_lost}. It has 12,031 training images. As there is no validation set for Pascal VOC dataset, therefore, we only evaluate FID on the train set. We sample 10,000 images and use 10,000 random-cropped 64-sized train images as reference set for FID evaluation.

\paragraph{COCO\_20K.}
We follow the split from~\citep{simeoni2021localizing_lost,vo2020toward,lin2014microsoft_mscoco}. COCO\_20k is a subset of the COCO2014 trainval dataset, consisting of 19,817 randomly chosen images, used in unsupervised object discovery~\citep{simeoni2021localizing_lost,vo2020toward}. We sample 10,000 images and use 10,000 random-cropped 64-sized train images as reference set for FID evaluation. %

\paragraph{COCO-Stuff.}
It has a train set of 49,629 images, validation set of  2,175 images, where the original classes are merged into 27 (15
stuff and 12 things) high-level categories. We use the dataset split following~\citep{hamilton2022unsupervised_stego, ji2019invariant_iic,cho2021picie,zhang2022dense_dsn}, 
We sample 10,000 images and use 10,000 train/validation images as reference set for FID evaluation.

\subsection{LOST, STEGO algorithms}
\label{sec:lost_stego_details}

\paragraph{LOST algorithm details.} 
We conduct padding to make the original image can be patchified to be fed into the \texttt{ViT} architecture~\citep{dosovitskiy2020image_vit}, and feed the original padded image into the \texttt{LOST} architecture using official source code~\footnote{https://github.com/valeoai/LOST}. \texttt{LOST} can also be utilized in a two-stage approach to provide multi-object, due to its complexity, we opt for only single-object discovery in this paper.

\paragraph{STEGO algorithm details.} 
We follow the official source code~\footnote{https://github.com/mhamilton723/STEGO}, and apply padding to make the original image can be fed into the \texttt{ViT} architecture to extract the self-segmented guidance signal.  

For COCO-Stuff dataset, we directly use the official pretrained weight. For Pascal VOC, we train \texttt{STEGO} ourselves using the official hyperparameters.

In \texttt{STEGO}'s pre-processing for the $k$-NN, the number of neighbors for $k$-NN is 7. The segmentation head  of \texttt{STEGO} is composed  of a  two-layer \texttt{MLP} (with \texttt{ReLU} activation) and outputs a 70-dimension feature. The learning rate is $5e-4$,  the batch size is 64.

\section{Qualitative results} 
\label{sec:more_qualitative}

\subsection{Assigning semantic descriptions in self-labeled/segmented guidance}

In order to control the semantic content of a sample using self-guidance we can assign descriptions to each self-supervised cluster by manually checking a few example images per cluster. This is much more scalable since the total number of training images available are multiple orders of magnitude greater than the number of clusters. Furthermore, images in the same self-supervised cluster are highly semantically coherent and humans can easily describe their shared abstract concept~\citep{laina2020quantifying}.

\begin{figure*}
    \centering
    \hspace*{0.25in}
\includegraphics[width=0.99\textwidth]{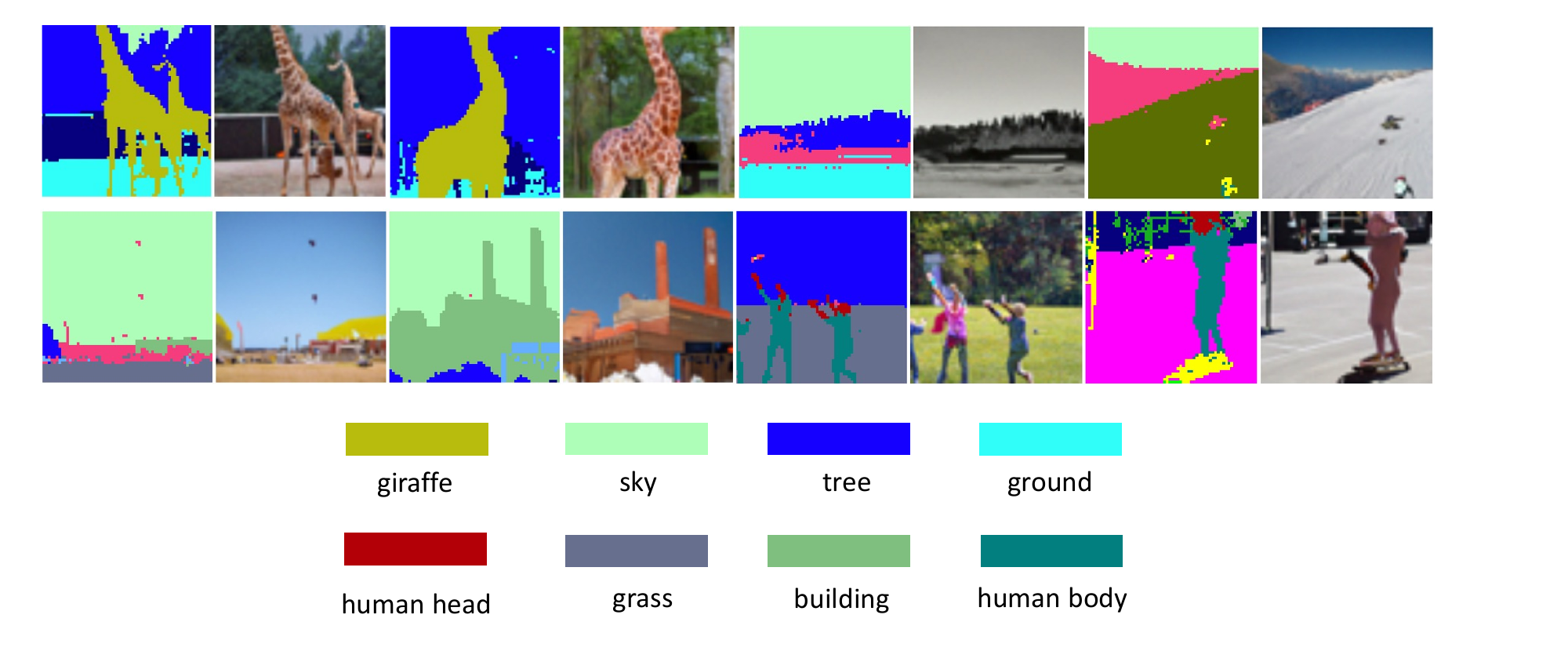}
    \caption{\textbf{Self-segmented guidance samples from COCO-Stuff companies with segmentation mask from \texttt{STEGO}~\citep{hamilton2022unsupervised_stego}.} The color map is shared among the overall dataset. The semantic description is deduced based on a few images. Best viewed in color.}
    \label{fig:cocostuff64_stego_with_semantics}
\end{figure*}

\begin{figure*}
    \centering
    \includegraphics[width=0.97\textwidth]{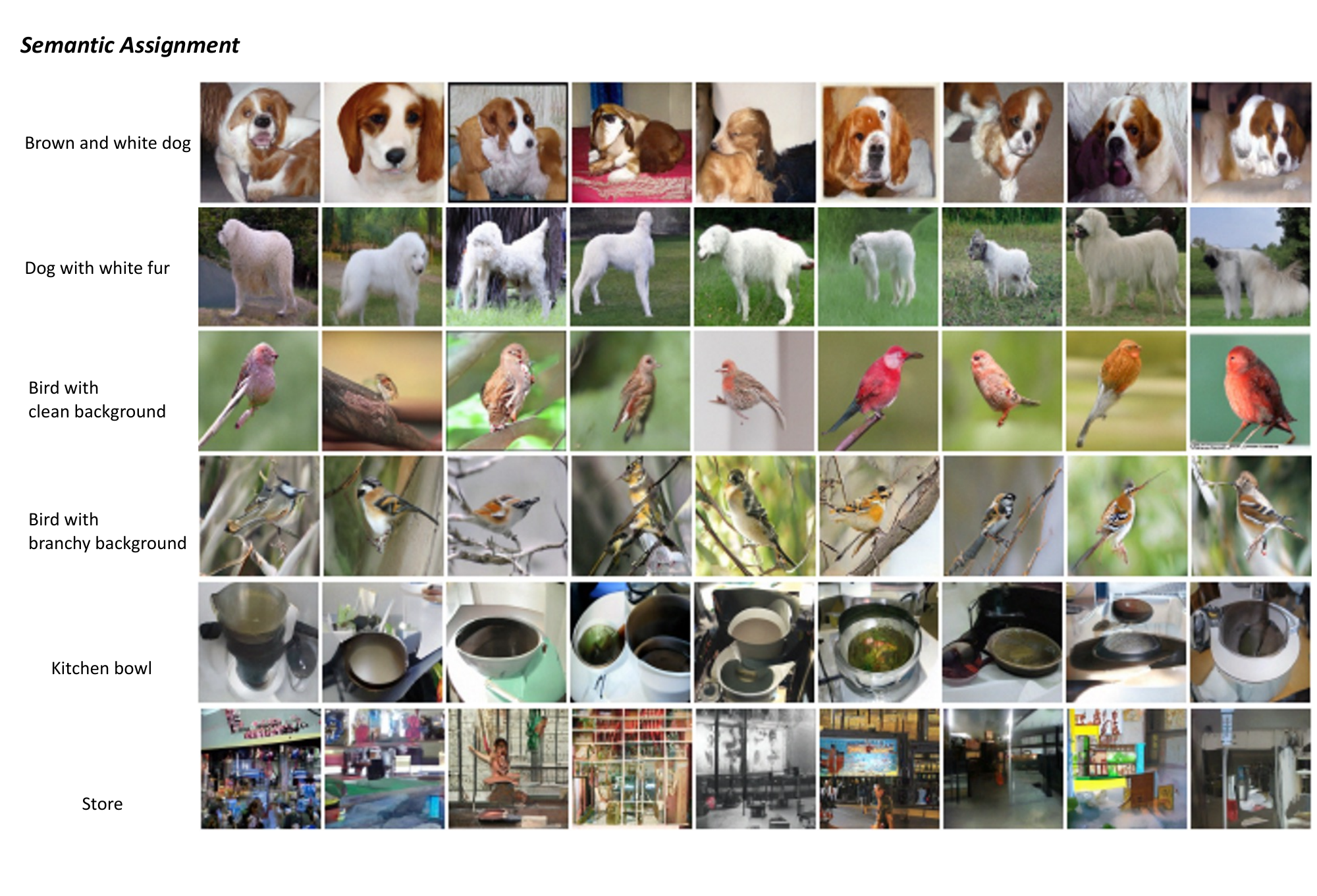}
    \caption{\textbf{Self-labeled guidance samples conditioning on the same guidance from ImageNet64.} We assign a cluster description based on a few sample images. Best viewed in color.}
    \label{fig:uncurated_in64_cluster_samecondition}
\end{figure*}

In~\Cref{fig:uncurated_in64_cluster_samecondition} we show examples of self-labeled guidance that highlight the semantic coherence of samples guided by the same cluster id. In~\Cref{fig:cocostuff64_stego_with_semantics} we show how this approach is also extendable to self-segmented guidance.

\subsection{More qualitative results}

\begin{figure*}
    \centering
    \includegraphics[width=0.99\textwidth]{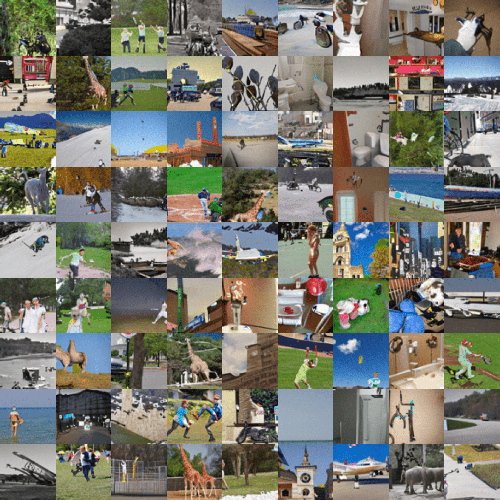}
    \caption{\textbf{Self-segmented guidance samples from COCO-Stuff.}  Best viewed in color.}
    \label{fig:uncurated_cocostuff64_stego_all}
\end{figure*}

\begin{figure*}
    \centering
    \includegraphics[width=0.99\textwidth]{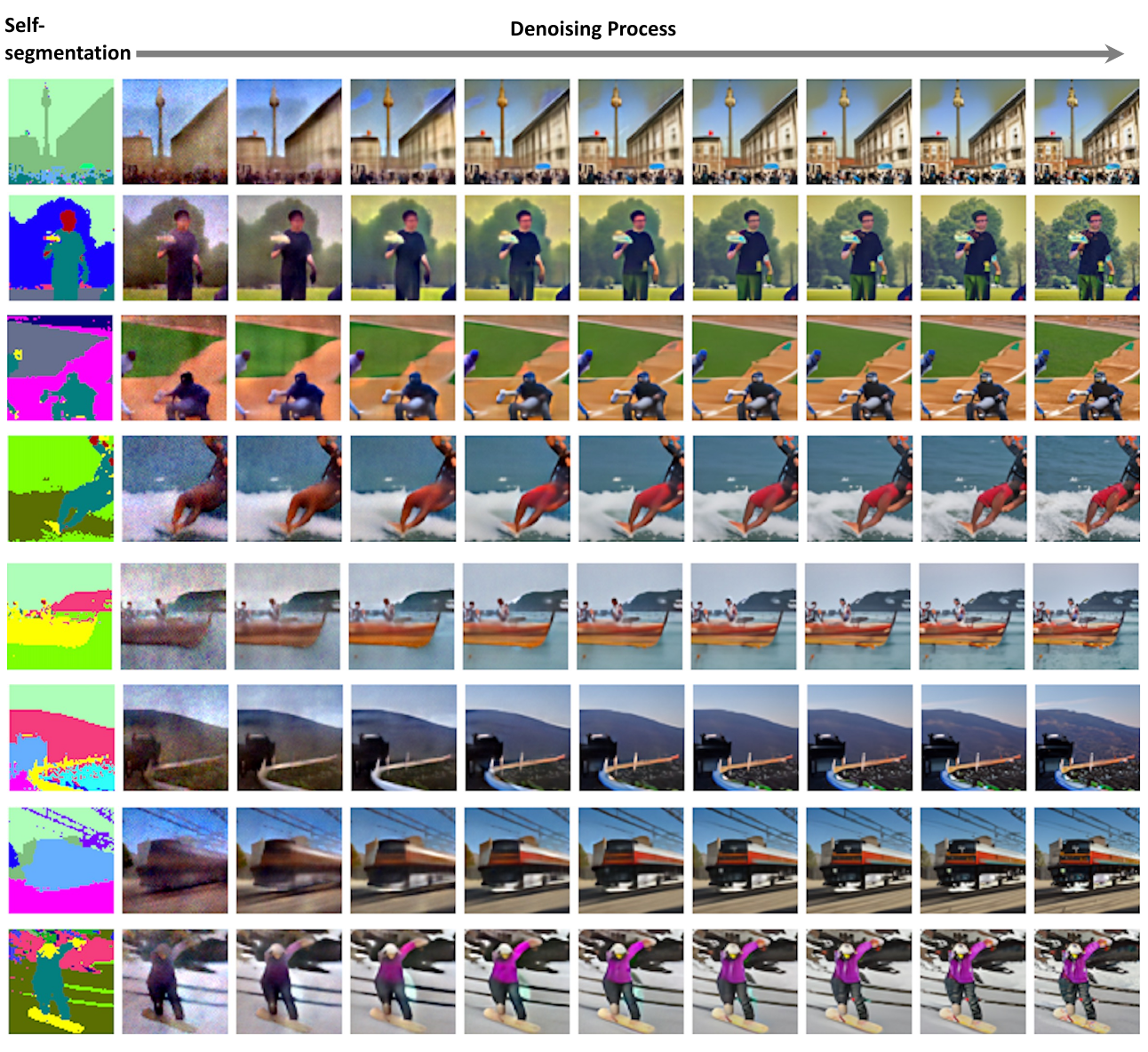}
    \caption{\textbf{Denoising process of self-segmented guidance samples (uncurated) from COCO-Stuff.}  The first column is the self-segmented guidance mask from \texttt{STEGO}~\citep{hamilton2022unsupervised_stego}, The remaining columns are from the noisiest period to the less noisy period. Best viewed in color.}
    \label{fig:chainvis_cocostuff_stego}
\end{figure*}

\begin{figure*}
    \centering
    \includegraphics[width=0.86\textwidth]{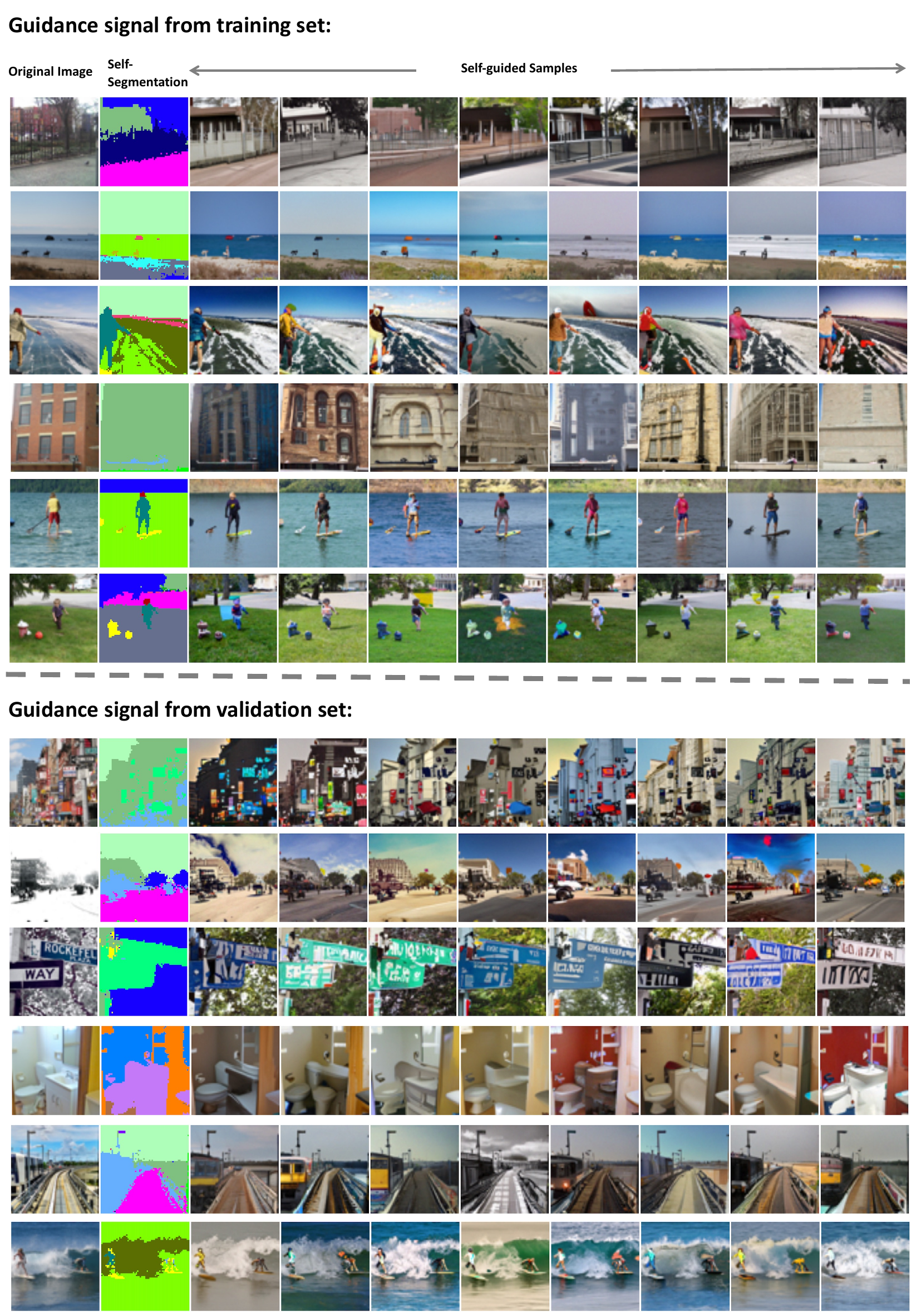}
    \caption{\textbf{Self-segmented guidance samples   (uncurated) from COCO-Stuff.}  The first column is the real image where we attain the conditional mask. The second column is the self-segmented mask we obtain from \texttt{STEGO}~\citep{hamilton2022unsupervised_stego}, The remaining columns are the random samples conditioning on the same self-segmented mask. Best viewed in color.}
    \label{fig:stego_vis_samecondition_cocostuff_more}
\end{figure*}

\begin{figure*}
    \centering
    \includegraphics[width=0.99\textwidth]{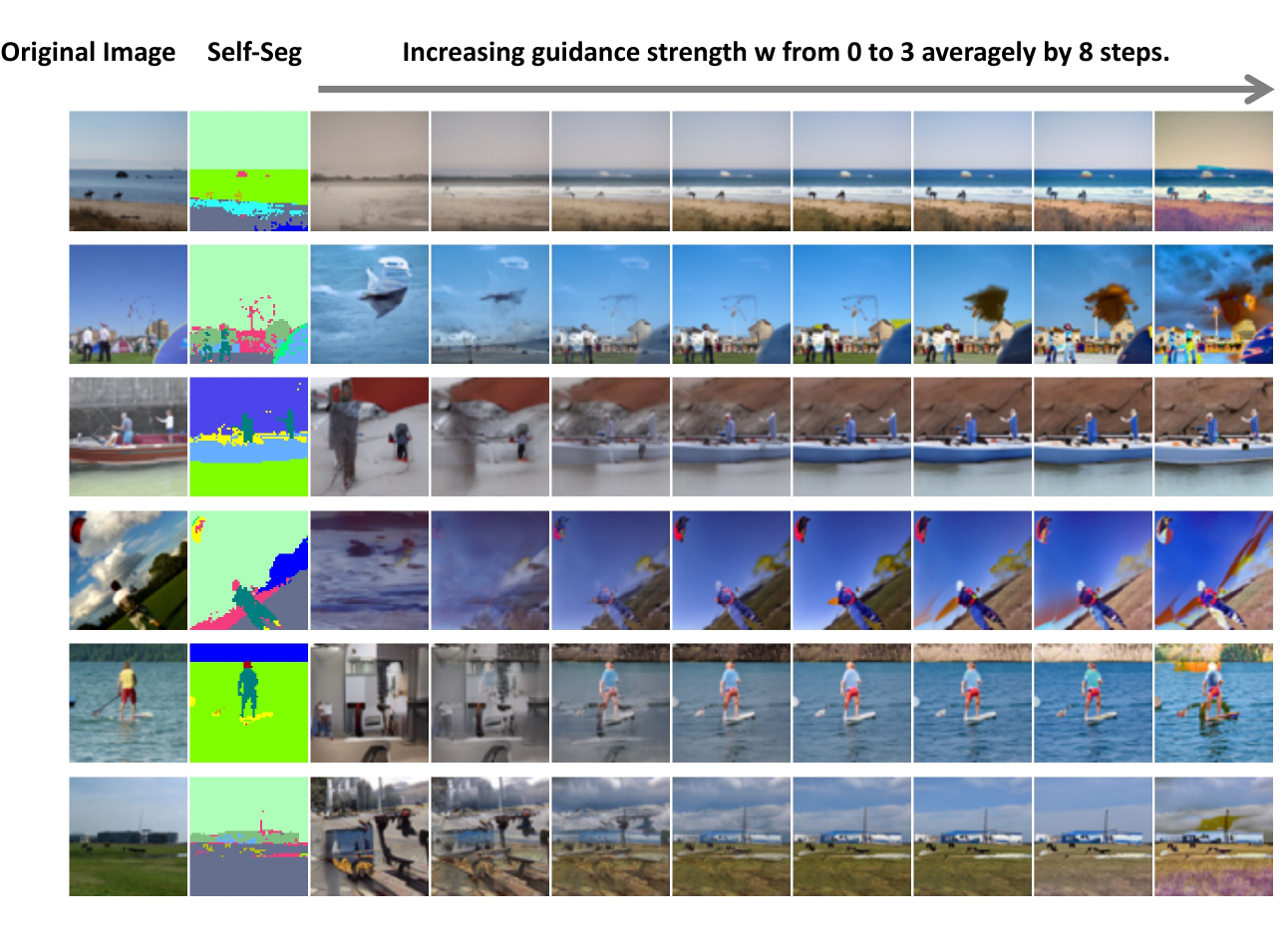}
    \caption{\textbf{Self-segmented guidance samples from Pascal VOC.}  The first column is the real image where we attain the conditional mask. The second column is the self-segmented mask we obtain from \texttt{STEGO}~\citep{hamilton2022unsupervised_stego}. The remaining columns are the visualization when we averagely increase guidance strength $w$ from 0 to 3 by 8 steps. Best viewed in color.}
    \label{fig:voc_stego_condscale0_3_step8}
\end{figure*}

\begin{figure*}
    \centering
    \includegraphics[width=0.89\textwidth]{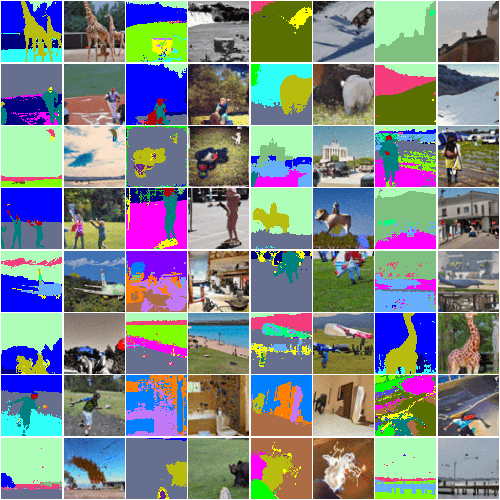}
    \caption{\textbf{Self-segmented guidance samples (uncurated)} from COCO-Stuff companies with segmentation mask from \texttt{STEGO}~\citep{hamilton2022unsupervised_stego}. The color map is shared among the overall dataset. Best viewed in color.}
    \label{fig:uncurated_cocostuff64_stego}
\end{figure*}

\begin{figure*}
    \centering
    \includegraphics[width=0.99\textwidth]{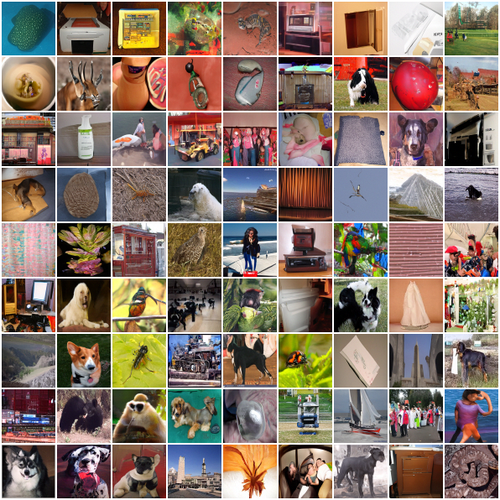}
    \caption{\textbf{Self-labeled guidance samples  (uncurated) from ImageNet64.}  Best viewed in color.}
    \label{fig:uncurated_in64}
\end{figure*}
\begin{figure*}
    \centering
    \includegraphics[width=0.99\textwidth]{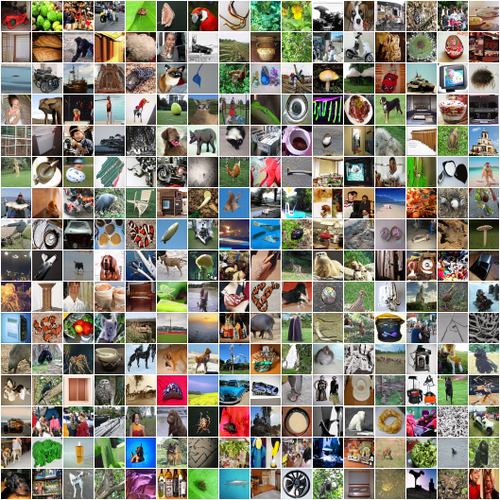}
    \caption{\textbf{Self-labeled guidance samples (uncurated) from ImageNet32.}  Best viewed in color.}
    \label{fig:uncurated_in32}
\end{figure*}

 \begin{figure*}
 \begin{subfigure}{.5\textwidth}
    \centering
    \includegraphics[width=0.99\textwidth]{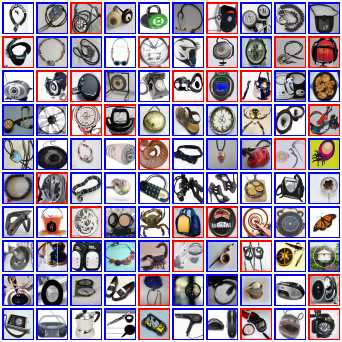}
    \caption{Querying by the sample in feature similarity. }
\end{subfigure}%
\begin{subfigure}{.5\textwidth}
    \centering
    \includegraphics[width=0.99\textwidth]{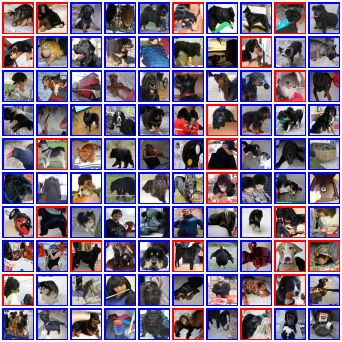}
    \caption{Querying by real images in feature similarity. }
\end{subfigure}
 \begin{subfigure}{.5\textwidth}
    \centering
    \includegraphics[width=0.99\textwidth]{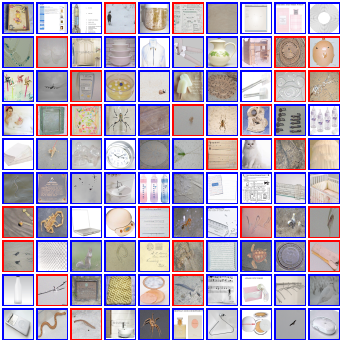}
    \caption{Querying by the sample in pixel similarity. }
\end{subfigure}%
\begin{subfigure}{.5\textwidth}
    \centering
    \includegraphics[width=0.99\textwidth]{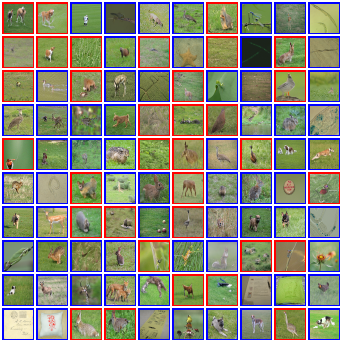}
    \caption{Querying by real images in pixel similarity. }
\end{subfigure}
\caption{\textbf{$k$-NN query result visualization.}
\textcolor{blue}{Blue} means samples, \textcolor{red}{red} means real images. Images are ordered from left to right, top to down, by SimCLR~\citep{chen2020big_simclr_v2} feature similarity or pixel similarity. Sampled images are sampled by DDIM~\citep{song2020denoising_ddim} with 250 steps. Guidance strength $w$ is 2. 
Firstly, we construct a gallery that is composed of an equivalent number of  sampled and real images, then we ablate two experiments by querying using sampled images or real images in feature space and image space. 
\textbf{Conclusion:}We can easily see, regardless of the feature space or image space, the $k$-NN query results are always highly semantic similar, and they show the diffusion model is not only to memorize the training data/real images but also can generalize well to synthesize novel images.}
\label{fig:knn_query_vis_in32}
\end{figure*}

\begin{figure*}
    \centering
    \includegraphics[width=0.99\textwidth]{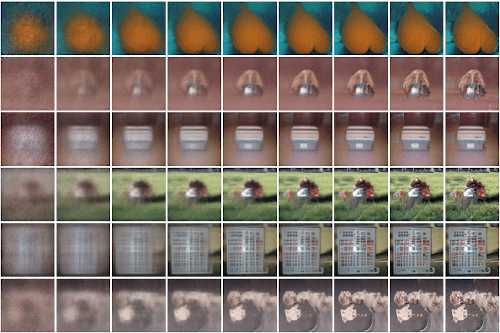}
    \caption{\textbf{Denoising process for ImageNet64.}
    }
    \label{fig:chainvis_in64}
\end{figure*}

\begin{figure*}
    \centering
    \includegraphics[width=0.99\textwidth]{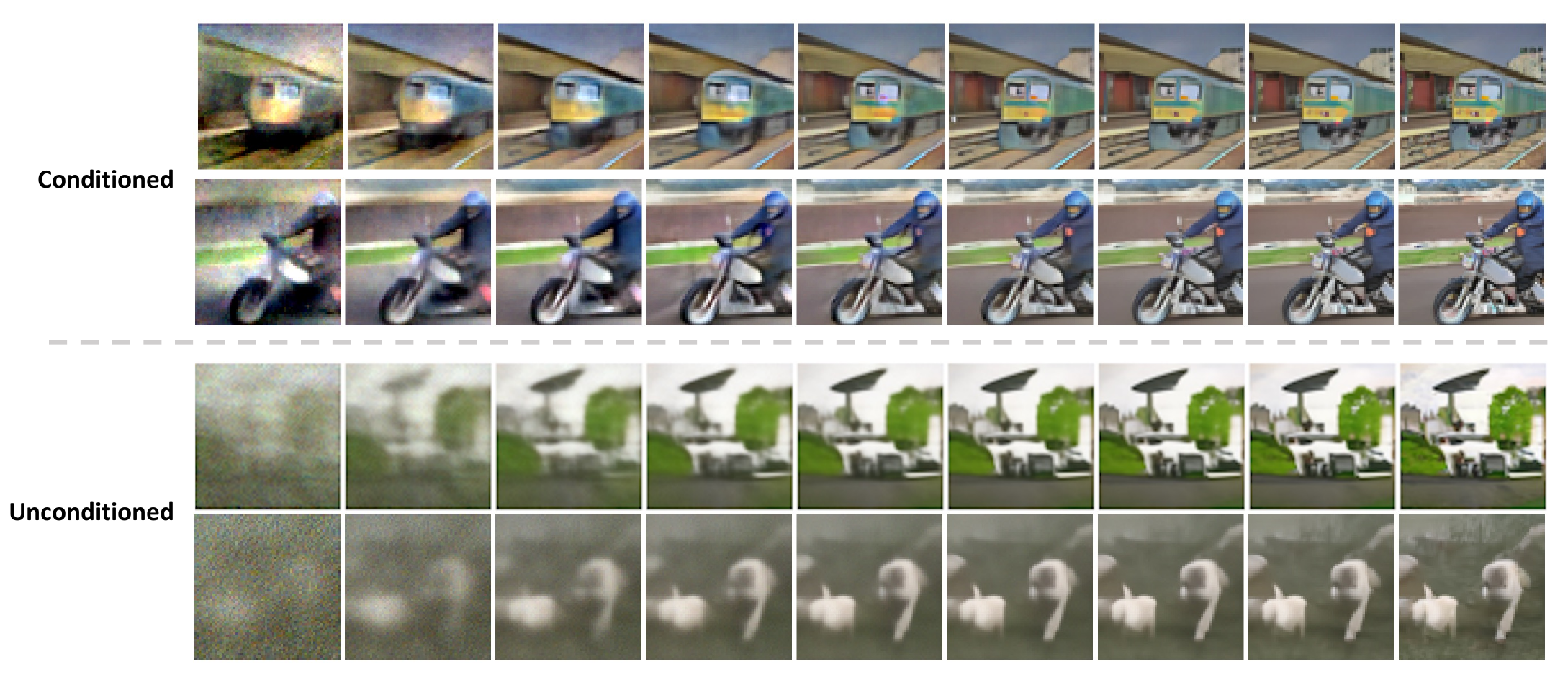}
    \caption{\textbf{Denoising process for Pascal VOC.} The first two rows are sampled from guidance strength $w=2$ using our self-segmented guidance, the last two rows are sampled from guidance strength $w=0$. By conditioning on our self-segmented guidance, the denoising process becomes easier and faster, this efficient denoising aligns with the observation from~\citep{preechakul2022diffusion_autoencoder}.}
    \label{fig:chainvis_voc}
\end{figure*}

\begin{figure*}
    \centering
    \includegraphics[width=0.99\textwidth]{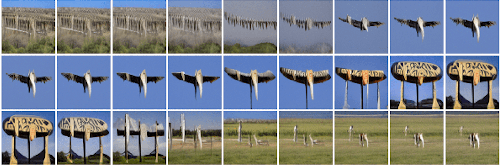}
    \caption{\textbf{Sphere interpolation between two random self-labeled guidance signals on ImageNet64.} The sphere interpolation follows the DDIM~\citep{song2020denoising_ddim}.  Best viewed in color.}
    \label{fig:interp_in64}
\end{figure*}

\begin{figure*}
\begin{subfigure}{.99\textwidth}
    \centering
    \includegraphics[width=0.99\textwidth]{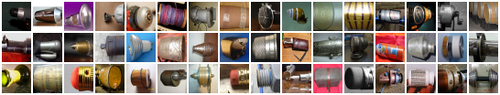}
    \caption{cluster625}
\end{subfigure}
\begin{subfigure}{.99\textwidth}
    \centering
    \includegraphics[width=0.99\textwidth]{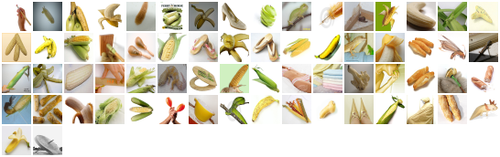}
    \caption{cluster807}
\end{subfigure}
\begin{subfigure}{.99\textwidth}
    \centering
    \includegraphics[width=0.99\textwidth]{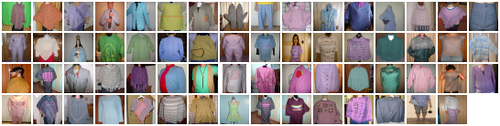}
    \caption{cluster890}
\end{subfigure}
\caption{\textbf{Cluster visualization of real images  in ImageNet32  after $k$-means.}
}
\label{fig:clustervis_in32}
\end{figure*}

\begin{figure*}
\begin{subfigure}{.5\textwidth}
    \centering
    \includegraphics[width=0.99\textwidth]{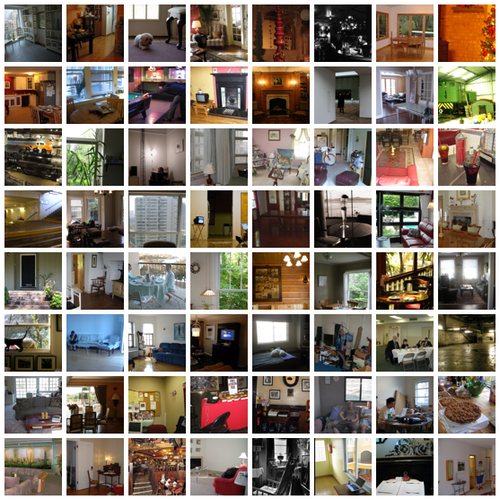}
    \caption{cluster17}
\end{subfigure}
\begin{subfigure}{.5\textwidth}
    \centering
    \includegraphics[width=0.99\textwidth]{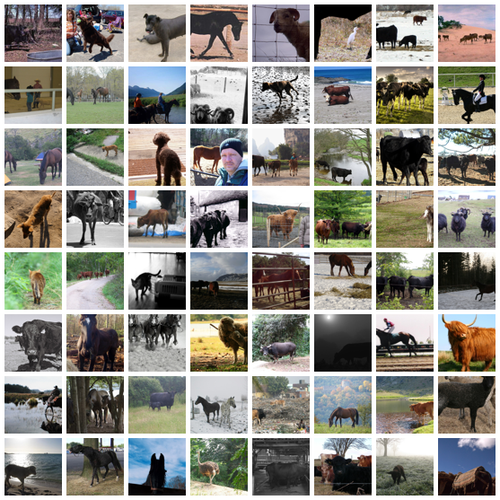}
    \caption{cluster18}
\end{subfigure}
\begin{subfigure}{.5\textwidth}
    \centering
    \includegraphics[width=0.99\textwidth]{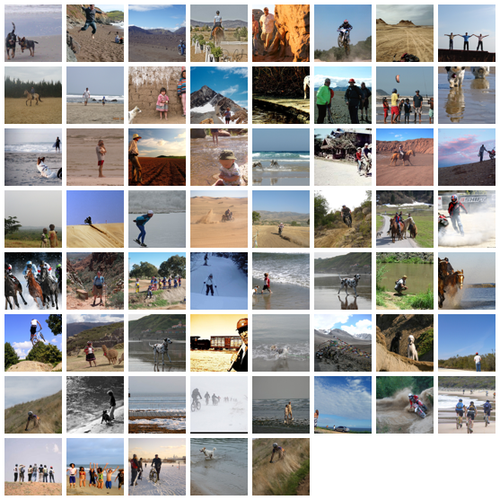}
    \caption{cluster45}
\end{subfigure}
\begin{subfigure}{.5\textwidth}
    \centering
    \includegraphics[width=0.99\textwidth]{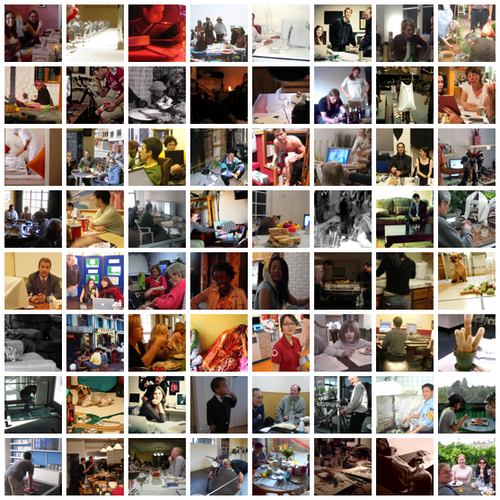}
    \caption{cluster50}
\end{subfigure}
\caption{\textbf{Cluster visualization of real images in Pascal VOC  after $k$-means.}
Best viewed by zooming in.}
\label{fig:clustervis_voc}
\end{figure*}

\begin{figure*}
    \centering
    \includegraphics[width=0.99\textwidth]{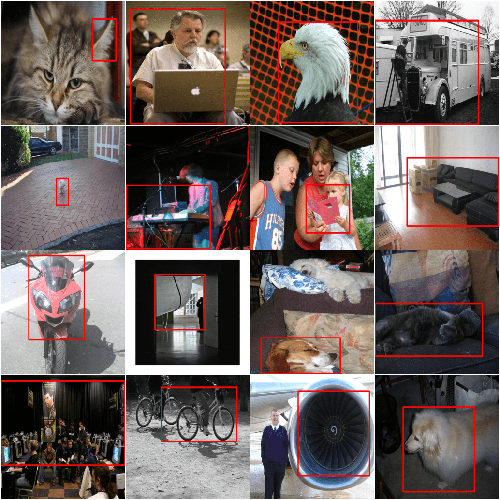}
    \caption{\textbf{Bounding box result from LOST on Pascal VOC}. As \texttt{LOST}~\citep{simeoni2021localizing_lost} is an unsupervised-learning method, some flaws in the generated box are expected. Images are resized squarely for better visualization.}
    \label{fig:lost_vis_voc}
\end{figure*}

\begin{figure*}
    \centering
    \includegraphics[width=0.99\textwidth]{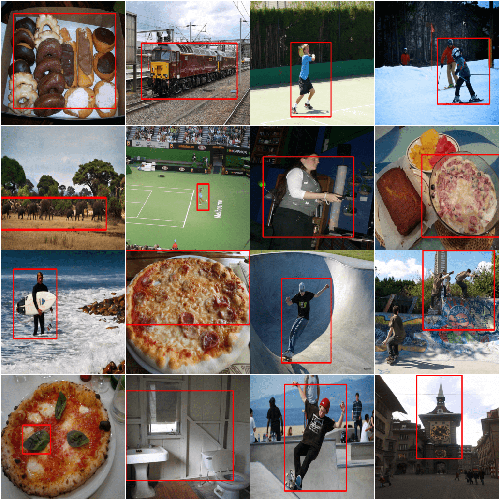}
    \caption{\textbf{Bounding box result from LOST~\citep{simeoni2021localizing_lost} on COCO\_20K.} Images are resized squarely for better visualization.}
    \label{fig:lost_vis_coco20k}
\end{figure*}

\begin{figure*}
    \centering
    \includegraphics[width=0.99\textwidth]{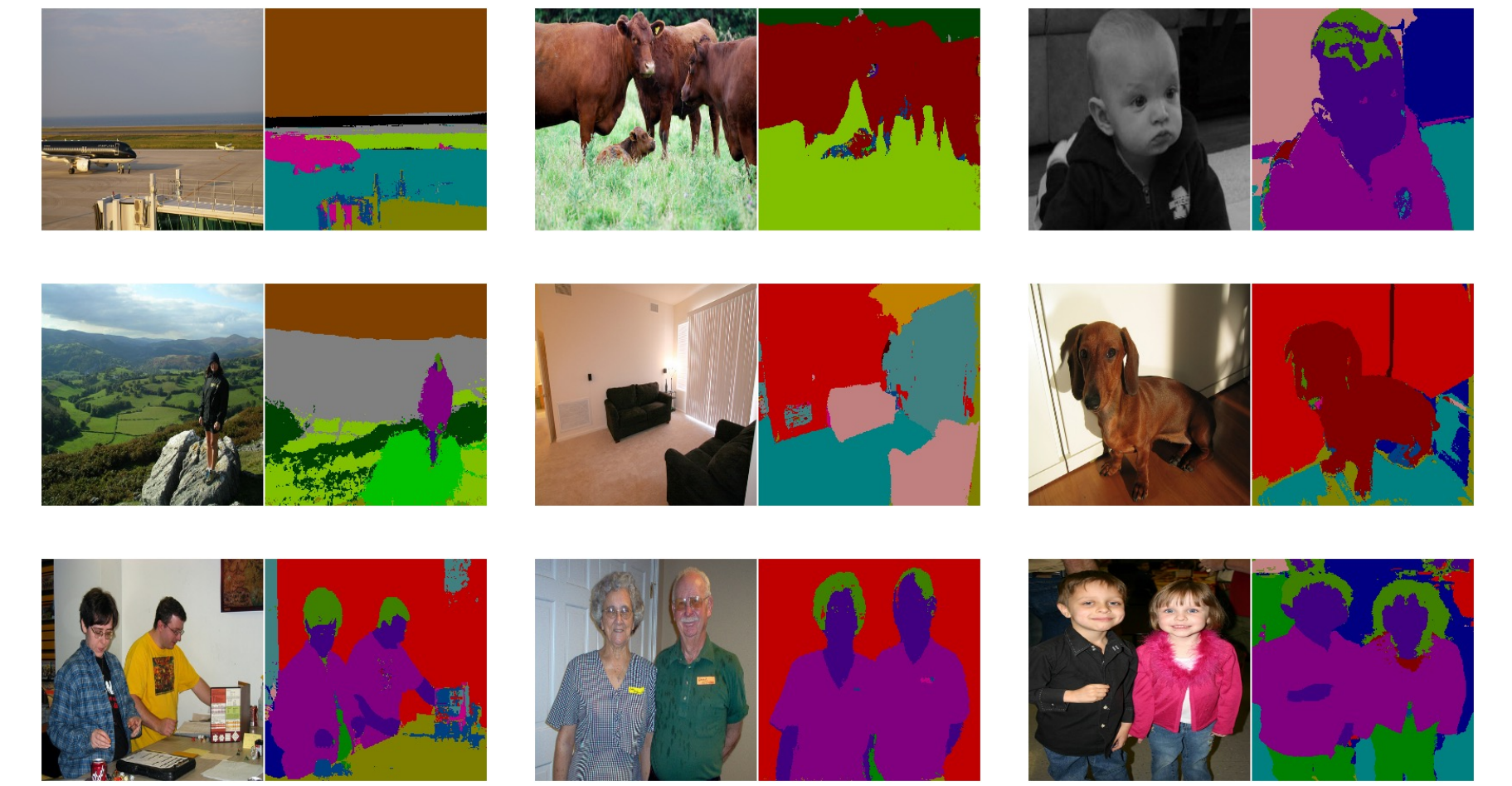}
    \caption{\textbf{Segmentation mask result from STEGO on Pascal VOC dataset}. Cluster number $k$ is 21.   Images are resized squarely for better visualization. The color map is shared among the overall dataset. Best viewed in color.}
    \label{fig:stego_vis_voc}
\end{figure*}

\end{document}